%% file: seqpaper-arxiv.tex
\RequirePackage{etoolbox}
\RequirePackage{etoolbox}
\csdef{input@path}{%
 {sty/}
 {img/}
}%
\csgdef{bibdir}{bib/}

\documentclass[ba]{imsart-arxiv}
\pubyear{\today}
\volume{}
\issue{}
\doi{}
\firstpage{1}
\lastpage{1}

\usepackage{bm}
\usepackage{dsfont}
\usepackage{amsthm}
\usepackage{amsmath}
\usepackage{amssymb}
\usepackage{amsfonts}
\usepackage{natbib}
\setcitestyle{notesep={ }}
\usepackage[colorlinks,citecolor=blue,urlcolor=blue,filecolor=blue,backref=page]{hyperref}
\usepackage{graphicx}
\usepackage[utf8]{inputenc}
\usepackage[T1]{fontenc}

\usepackage{algorithm}
\usepackage{algpseudocode}

\DeclareMathOperator*{\argmax}{arg\,max}

\startlocaldefs
\numberwithin{equation}{section}
\theoremstyle{plain}

\endlocaldefs

\usepackage{xcolor}

\newcommand{\thetab}{\bm{\theta}}
\newcommand{\dbf}{\mathbf{d}}
\newcommand{\ybf}{\mathbf{y}}

\begin{document}

\begin{frontmatter}
\title{Sequential Bayesian Experimental Design for Implicit Models via Mutual Information} 
\runtitle{Sequential Bayesian Experimental Design for Implicit Models}

\begin{aug}
\author{\fnms{Steven} \snm{Kleinegesse}\thanksref{addr1}\ead[label=e1]{steven.kleinegesse@ed.ac.uk}}, 
\author{\fnms{Christopher} \snm{Drovandi}\thanksref{addr2}\ead[label=e2]{c.drovandi@qut.edu.au}}
\and
\author{\fnms{Michael U.} \snm{Gutmann}\thanksref{addr1}\ead[label=e3]{michael.gutmann@ed.ac.uk}}

\runauthor{S. Kleinegesse, C. Drovandi and M. Gutmann}

\address[addr1]{University of Edinburgh,
    \printead{e1}, 
    \printead*{e3}
}

\address[addr2]{Queensland University of Technology,
    \printead{e2}
}


\end{aug}

\begin{abstract}
Bayesian experimental design (BED) is a framework that uses
statistical models and decision making under uncertainty to optimise
the cost and performance of a scientific experiment. Sequential BED,
as opposed to static BED, considers the scenario where we can
sequentially update our beliefs about the model parameters through
data gathered in the experiment. A class of models of particular
interest for the natural and medical sciences are implicit models,
where the data generating distribution is intractable, but sampling
from it is possible. Even though there has been a lot of work on
static BED for implicit models in the past few years, the notoriously
difficult problem of sequential BED for implicit models has barely
been touched upon. We address this gap in the literature by devising a
novel sequential design framework for parameter estimation that uses
the Mutual Information (MI) between model parameters and simulated
data as a utility function to find optimal experimental designs, which
has not been done before for implicit models. Our approach uses
likelihood-free inference by ratio estimation to simultaneously
estimate posterior distributions and the MI. During the sequential BED
procedure we utilise Bayesian optimisation to help us optimise the MI
utility. We find that our framework is efficient for the various
implicit models tested, yielding accurate parameter estimates after
only a few iterations.


\end{abstract}

\begin{keyword}[class=MSC]
\kwd[Primary ]{62K05}
\kwd[; secondary ]{62L05}
\end{keyword}

\begin{keyword}
\kwd{Bayesian Experimental Design}
\kwd{Likelihood-Free Inference}
\kwd{Mutual Information}
\kwd{Approximate Bayesian Computation}
\kwd{Implicit Models}
\end{keyword}

\end{frontmatter}

\input{main}

\clearpage


\appendix
\input{supplementary}
\clearpage
\bibliographystyle{ba}
\bibliography{references}

\begin{acknowledgement}
Steven Kleinegesse was supported in part by the EPSRC Centre for Doctoral Training in Data Science, funded by the UK Engineering and Physical Sciences Research Council (grant EP/L016427/1) and the University of Edinburgh. Christopher Drovandi was supported by an Australian Research Council Discovery Project (DP200102101).
\end{acknowledgement}

\end{document}

%% file: main.tex
\section{Introduction}
Scientific experiments are critical to improving our perception and understanding of how the world works. Most of the time these experiments are time-consuming and expensive to perform. It is thus crucial to decide where and how to collect the necessary data to learn most about the subject of study. \emph{Bayesian experimental design} attempts to solve this problem by allocating resources in an experiment using Bayesian statistics (see~\citet{Ryan2016} for a comprehensive review). Roughly speaking, the aim is to find experimental design, e.g.~measurement location or time, that are expected to most rapidly address the scientific aims of the experiment, mitigating the costs. The relevant scientific objectives can include, but are not limited to, model parameter estimation, prediction of future observations or comparison of competing models. In this particular paper we shall only be concerned with the objective of parameter estimation.

At the core of Bayesian experimental design is the so-called utility function, which is maximised to find the optimal design at which to perform an experiment. A popular and principled utility function for parameter estimation is the \emph{mutual information} between model parameters and simulated data~\citep{Lindley1972}. Intuitively, this metric measures the additional information we would obtain about the model parameters given some real-world observations taken at a particular design. Depending on the model, computing the mutual information can be difficult or even intractable and, as a consequence, various methods for its estimation have arisen.

Whenever new, real-world data is collected through physical experiments, the surface of the utility function tends to change, e.g~collecting data with the same design would generally not yield much new information. The treatment of this change, for a single, new data point, is called myopic \emph{sequential} Bayesian experimental design and is manifested through an update of the prior distribution upon observing real-world data.
This stands in contrast to \emph{static} Bayesian experimental design that is concerned with situations where we do not update our prior distributions when observing new data, such as when there is nearly no time, or too much time, between real-world measurements, or data has to be collected all at once. Sequential Bayesian experimental design is a well-established field for situations in which the model has a tractable likelihood function and inferring the posterior distribution is straight-forward~\citep{Ryan2016}. However, there have only been few studies~\citep[e.g.][]{Hainy2016b} pertaining to the arguably more realistic situation of intractable, \emph{implicit models}.

In practice, statistical models commonly have likelihood functions
that are analytically unknown or intractable. This is the case for
implicit models, where we cannot evaluate the likelihood but we can
still sample from it. They are ubiquitous in the natural and medical
sciences and therefore have widespread use. Examples include ecology
\citep{Ricker1954, Wood2010}, epidemiology~\citep{Numminen2013,
  Corander2017}, genetics~\citep{Marttinen2015, Arnold2018},
cosmology~\citep{Schafer2012, Alsing2018} and modelling particle
collisions~\citep{Agostinelli2003, Sjostrand2008}. Because the
likelihood function for implicit models is intractable, we are
generally not able to work with the exact posterior distribution. As a
result, likelihood-free inference methods have emerged to solve this
issue.

In order to compute the mutual information between model parameters
and simulated data however, one needs to be able to evaluate the ratio
between posterior density to prior density several times which is
difficult in the likelihood-free setting. This is especially
challenging in the sequential framework, where the current belief
distribution gets updated after every observation. In this work we
propose to approximate the density ratio in mutual information
directly via the Likelihood-Free Inference by Ratio Estimation
(LFIRE) method of \citet{Thomas2016}. We perform this in the context of
sequential Bayesian experimental design, a significant extension
of~\citet{Kleinegesse2019} that only considered the static setting.

In this paper we propose a sequential Bayesian experimental design
framework for implicit models that have intractable data-generating
distributions. In brief, we make the following contributions:

\begin{enumerate}
  \item Our approach allows us to approximate the mutual information in the presence of an implicit model directly by LFIRE, without resorting to simulation-based likelihood approximations required by other approaches.  At the same time, LFIRE also provides an approximation of the sequential posterior.
  \item We demonstrate the efficacy of our sequential framework on examples from epidemiology and cell biology. We further
    showcase that previous approaches may produce experimental designs
    that heavily penalise multi-modal posteriors thereby introducing
    an undesirable bias into the scientific data gathering stage,
    which our approach avoids.
\end{enumerate}

In Section~\ref{sec:background} we give basic background knowledge to
sequential Bayesian experimental design, mutual information and
likelihood-free inference, in particular LFIRE. We then combine these
concepts in Section~\ref{sec:framework} and explain our novel
framework of sequential design for implicit models. We test our
framework on various implicit models and present the results in
Section~\ref{sec:experiments}. We conclude our work and discuss
possible future work in Section~\ref{sec:conclusion}.

\section{Background} \label{sec:background}

\subsection{Bayesian Experimental Design}

In Bayesian experimental design the aim is to find experimental designs $\dbf$ that yield more informative, or useful, real-world observations than others. Furthermore, in this work we are particularly interested in finding the optimal design $\dbf^\ast$ that results in the best estimation of the model parameters. At its core, this task requires defining a utility function $U(\dbf)$ that describes the value of performing an experiment at $\dbf \in \mathcal{D}$, where $\mathcal{D}$ defines the space of possible designs. In order to qualify as a `fully Bayesian design', this utility has to be a functional of the posterior distribution $p(\thetab \mid \dbf, \ybf)$~\citep{Ryan2016}, where $\thetab$ are the model parameters and $\ybf$ is simulated data. The utility function is then maximised in order to find the optimal design $\dbf^\ast$, i.e.
\begin{equation} \label{eq:optdesign}
\dbf^\ast = \argmax_{\dbf \in \mathcal{D}} U(\dbf).
\end{equation}
The choice of utility function $U(\dbf)$ is thus critical, as different functions will usually lead to different optimal designs. The most suitable utilities naturally depend on the task in question, but there are a few common functions that have been used extensively in the literature. For instance, the \emph{Bayesian D-Optimality} (BD-Opt) is based on the determinant of the inverse covariance matrix of the posterior distribution,\footnote{See the Appendix~\ref{app:A} for an alternative form of the BD-Opt utility.} and is a measure of how precise the resulting posterior might be given certain designs~\citep{Ryan2016},
\begin{equation} \label{eq:bdopt}
U(\dbf) = \mathbb{E}_{p(\ybf \mid \dbf)}\left[\frac{1}{\text{det}(\text{cov}(\thetab \mid \ybf, \dbf))}\right]. 
\end{equation}
While BD-Opt works well for uni-modal posteriors, it is not suitable for multi-modal or complex posteriors as it heavily penalises diversion from uni-modality. A more versatile and robust utility function is the \emph{mutual information}, one of the most principled choices in Bayesian experimental design~\citep[e.g.][]{Ryan2016}.


\subsection{Mutual Information}

The mutual information $\mathrm{I}(\thetab; \ybf | \dbf)$ can be interpreted as the expected reduction in uncertainty (entropy) of the model parameters if the data $\ybf$ was obtained with design $\dbf$. It accounts for possibly non-linear dependencies between $\thetab$ and $\ybf$.
It is an effective metric with regards to the task of parameter estimation, as we are essentially concerned with finding the design for which the corresponding observation yields the most information about the model parameters $\thetab$. In other words, mutual information tells us how `much' we can learn about the model parameters given the prospective data at a particular design.

Mutual information is defined as the Kullback-Leibler (KL) divergence $\mathrm{D_{KL}}$~\citep{Kullback1951} between the joint distribution and the product of marginal distributions of $\ybf$ and $\thetab$ given $\dbf$, i.e.
\begin{align}
  \mathrm{I}(\thetab; \ybf \mid \dbf) &= \mathrm{D_{KL}}(p(\thetab, \ybf \mid \dbf) \mid\mid p(\bm{\theta \mid \dbf})p(\ybf \mid \dbf)) \\
  &= \mathrm{D_{KL}}(p(\thetab, \ybf \mid \dbf) \mid\mid p(\thetab)p(\ybf \mid \dbf)) \label{eq:mutual_origin1} \\
&= \int p(\thetab, \ybf \mid \dbf) \log\left[\frac{p(\thetab, \ybf \mid \dbf)}{p(\thetab)p(\ybf \mid \dbf)}\right] \mathrm{d}\thetab \mathrm{d}\ybf, \label{eq:mutual_origin2}
\end{align}
where we have made the usual assumption that our prior belief about $\thetab$ is not affected by the design, i.e.~$p(\thetab \mid \dbf)$ = $p(\thetab)$. 

The mutual information can also be interpreted as the expected KL divergence between posterior $p(\thetab \mid \dbf, \ybf)$ and prior $p(\thetab)$~\citep[see e.g.][]{Ryan2016} and essentially tells us how different on average our posterior distribution is to the prior distribution. The utility that we then need to maximise in order to find the optimal design $\dbf^\ast$ is thus
\begin{align}
U(\dbf) &= \mathrm{I}(\thetab; \ybf \mid \dbf) \label{eq:mutual1} \\
&= \mathbb{E}_{p(\ybf \mid \dbf)}[\mathrm{D_{KL}}(p(\thetab \!\mid\! \dbf, \ybf) \mid\mid p(\thetab))] \label{eq:mutual2} \\
&= \int \log\left[\frac{p(\thetab \!\mid\! \dbf, \ybf)}{p(\thetab)}\right] p(\thetab) p(\ybf \!\mid\! \thetab, \dbf) \mathrm{d}\thetab \mathrm{d}\ybf, \label{eq:mutual3}
\end{align}
where $p(\ybf \!\mid\! \thetab, \dbf)$ is the data generating
distribution, commonly referred to as the likelihood. The particular
form of mutual information in~\eqref{eq:mutual3} can be obtained from
\eqref{eq:mutual_origin2} by applying the product rule to $p(\thetab,
\ybf \!\mid\! \dbf)$. 
Even though mutual information is a well-studied concept, estimating it
efficiently remains an open question, especially in higher dimensions.

Assuming for now that we have optimised the utility function
in~\eqref{eq:mutual3} and have obtained the optimal design
$\dbf^\ast$. An experimenter would then go and perform the experiment
at $\dbf^\ast$ and observe real-world data $\ybf^\ast$. Everything up
to this point is $\emph{static}$ Bayesian experimental design. If we
would like to update our optimal design in light of a real-world
observation, we would have to perform $\emph{sequential}$ Bayesian
experimental design, i.e.~update our prior distribution and optimise
the utility function again. This procedure is then repeated several
times to obtain (myopic) sequentially designed experiments. We shall
not aim to find non-myopic sequentially designed experiments where we
would we plan ahead more than one time-step as this adds another layer
of complexity.
%

Let $k$ be the $k$th iteration of the sequential design procedure,
where $k=1$ corresponds to the task of finding the first optimal
experimental design $\dbf^\ast_1$ yielding real-world observation
$\ybf^\ast_1$. At iteration $k$ we then optimise the utility function
$U_k(\dbf)$ to obtain sequential optimal designs $\dbf_k^\ast$, with
corresponding real-world observations $\ybf_k^\ast$. The utility
function at iteration $k \in \{1, 2, \ldots, K\}$ depends on the set
of all previous observations $\mathbb{D}_{k-1} = \{\dbf^\ast_{1:k-1},
\ybf^\ast_{1:k-1}\}$, with $\mathbb{D}_0 = \varnothing$, and therefore
will change at every iteration. Its form stays similar
to~\eqref{eq:mutual3}, except that the prior and posterior
distributions now depend on $\mathbb{D}_{k-1}$, i.e.
\begin{align}
U_k(\dbf) &= \int \log\left(\frac{p(\thetab \mid \dbf, \ybf, \mathbb{D}_{k-1})}{p(\thetab \mid \mathbb{D}_{k-1})}\right) p(\ybf \mid \thetab, \dbf) p(\thetab \mid \mathbb{D}_{k-1}) \mathrm{d}\thetab \mathrm{d}\ybf. \label{eq:sequtility} 
\end{align}
Note that we will assume that data is generated independently of previous observations, i.e.~$p(\ybf \mid \thetab, \dbf, \mathbb{D}_{k-1}) = p(\ybf \mid \thetab, \dbf)$. In certain special cases where gathering real-world observations changes the data-generating process this would not be the case. 

\subsection{Likelihood-Free Inference} \label{sec:lfire}

Implicit models have intractable likelihood functions, which means that $p(\ybf \mid \thetab, \dbf)$ is either too expensive to compute or there is no closed-form expression. This results in standard Bayesian inference becoming infeasible. Because of their widespread use however, it is crucial to be able to infer the parameters of implicit models. As a result, the field of \emph{likelihood-free inference} has emerged. These methods leverage the fact that, by definition, implicit models allow for sampling from the data-generating distribution.

A popular likelihood-free approach is Approximate Bayesian
Computation~\citep[ABC,][]{Rubin1984}. ABC rejection
sampling~\citep{Pritchard1999}, the simplest form of ABC, works by
generating samples from the prior distribution over the model
parameters and then using them to simulate data from the implicit
model. The prior parameters that result in data that is `close' to
observed data are then accepted as samples from the ABC posterior
distribution. See~\citet{Sisson2018} or \citet{Lintusaari2017} for
reviews on ABC.

Since standard ABC is notoriously slow and requires tuning of some
hyperparameters, there has been considerable research in making
likelihood-free inference more efficient, using, for example, ideas from
Bayesian optimisation and experimental design \citep{Gutmann2016a,
  Jarvenpaa2019a, Jarvenpaa2020}, conditional density estimation
\citep{Papamakarios2016, Lueckmann2017, Greenberg2019}, classification
\citep{Gutmann2018}, indirect inference \citep{Drovandi2015},
optimisation \citep{Meeds2015, Ikonomov2020}, and more broadly
surrogate modelling with Gaussian processes \citep{Wilkinson2014,
  Meeds2015} and neural networks \citep{Blum2010, Chen2019,
  Papamakarios2019}.



In this paper, we make use of another approach to
likelihood-free inference called Likelihood-Free Inference by Ratio
Estimation (LFIRE)~\citep{Thomas2016}. LFIRE uses density ratio
estimation to obtain ratios $r(\dbf, \ybf, \thetab)$ of the likelihood
to marginal density and, therefore, the posterior to prior density,
i.e.
\begin{equation} \label{eq:ratio}
r(\dbf, \ybf, \thetab) = \frac{p(\ybf \mid \thetab, \dbf)}{p(\ybf \mid \dbf)} = \frac{p(\thetab \mid \dbf, \ybf)}{p(\thetab)}.
\end{equation}
The method works by estimating the ratio from data simulated from the
likelihood $p(\ybf \mid \thetab, \dbf)$ and data simulated from the
marginal $p(\ybf \mid \dbf)$, e.g.\ via logistic regression
\citep{Thomas2016}. Since the prior density $p(\thetab)$ is known,
learning the ratio corresponds to learning the posterior,
i.e.\ $\widehat{p}(\thetab \mid \dbf, \ybf) = \widehat{r}(\dbf, \ybf,
\thetab) p(\thetab)$. Importantly, the learned ratio yields
automatically also an estimate of the mutual information
in~\eqref{eq:sequtility}.

The LFIRE framework can be used with arbitrary models of the ratio
or posterior. For simplicity, like in the simulations by
\citet{Thomas2016}, we here use the log-linear model
\begin{equation} \label{eq:lfire_ratio}
\widehat{r}(\dbf, \ybf, \thetab) = \exp \left( \bm{\beta}(\dbf, \thetab)^\top \bm{\psi}(\ybf) \right),
\end{equation}
where $\bm{\psi}(\ybf)$ are some fixed summary
statistics. \citet{Thomas2016} showed that this log-linear model,
while simple, generalises the popular synthetic likelihood approach by
\citet{Wood2010, Price2018b}. Moreover, learning the summary
statistics from data, e.g.\ by means of neural networks, is possible
too \citep{Dinev2018}. For further details on LFIRE, we refer the
reader to the original paper by \citet{Thomas2016}.

\section{Sequential Mutual Information Estimation} \label{sec:framework}

The main aim of this work is to construct an effective sequential
experimental design framework for implicit models. To do this, we have
to approximate the sequential utility in~\eqref{eq:sequtility} in a
tractable manner. We propose to use LFIRE to estimate the intractable
density ratio in~\eqref{eq:sequtility} and, at the same time, obtain
the posterior density.
The main difference to the work of~\citet{Kleinegesse2019} is that
they only considered static experimental design and did not have the
additional complications that come with the sequential setting, such
as updating the prior distribution upon observing real-world data. Our
approach bears some similarities to the SMC sequential design method
of~\citet{Hainy2016b}. However, we use LFIRE for
updating the posterior when new data are collected and for direct
estimation of the mutual information (MI), rather than relying on
simulation-based likelihood estimation. Further,
unlike~\citet{Hainy2016b}, our approach avoids the MCMC perturbation
step, which requires re-processing all data seen so far.


\subsection{Sequential Utility}

We assume that we have already made $k-1$ experiments resulting in the set of optimal designs and observations $\mathbb{D}_{k-1} = \{\dbf^\ast_{1:k-1}, \ybf^\ast_{1:k-1}\}$, with $\mathbb{D}_0 = \varnothing$. At iteration $k$ of the sequential BED procedure we then set out to determine the optimal design $\dbf^\ast_k$ and the corresponding real-world observation $\ybf^\ast_k$. To do so, we first approximate the density ratio of $p(\thetab \mid \dbf, \ybf, \mathbb{D}_{k-1})$ and $p(\thetab \mid \mathbb{D}_{k-1})$ by the ratio $\widehat{r}_{k}(\dbf, \ybf, \thetab, \mathbb{D}_{k-1})$ computed by LFIRE,\footnote{Note that LFIRE actually estimates the log ratio of posterior to prior density.} such that
%
\begin{equation}
\widehat{r}_{k}(\dbf, \ybf, \thetab, \mathbb{D}_{k-1}) \approx \frac{p(\thetab \mid \dbf, \ybf, \mathbb{D}_{k-1})}{p(\thetab \mid \mathbb{D}_{k-1})}. \label{eq:seqratio}
\end{equation}
%
We then plug this into the expression for the sequential MI utility in~\eqref{eq:sequtility} and obtain
\begin{align}
U_k(\dbf) &= \int \log\left(\frac{p(\thetab \mid \dbf, \ybf, \mathbb{D}_{k-1})}{p(\thetab \mid \mathbb{D}_{k-1})}\right) p(\ybf \mid \thetab, \dbf) p(\thetab \mid \mathbb{D}_{k-1}) \mathrm{d}\thetab \mathrm{d}\ybf \\
&\approx \int \log\left(\widehat{r}_{k}(\dbf, \ybf, \thetab, \mathbb{D}_{k-1})\right) p(\ybf \mid \thetab, \dbf) p(\thetab \mid \mathbb{D}_{k-1}) \mathrm{d}\thetab \mathrm{d}\ybf. \label{eq:sequtility_implicit}
\end{align}
%
We can approximate this with a Monte-Carlo sample average to obtain the estimate
\begin{align}
\widehat{U}_k(\dbf) =& \frac{1}{N} \sum_{i=1}^N \log\left[\widehat{r}_k(\dbf, \ybf^{(i)}, \thetab^{(i)}, \mathbb{D}_{k-1})\right], \label{eq:sequtility_mc}
\end{align}
where $\ybf^{(i)} \sim p(\ybf \mid \dbf, \thetab^{(i)})$ and $\thetab^{(i)} \sim p(\thetab \mid \mathbb{D}_{k-1})$. The above mutual information estimate $\widehat{U}_k(\dbf)$ is then optimised to find the optimal design $\dbf_k^\ast$ and, through a real-world experiment, the corresponding observation $\ybf_k^\ast$ at iteration $k$.

Two core technical difficulties in~\eqref{eq:sequtility_mc} are (1) how to obtain parameter samples $\thetab^{(i)} \sim p(\thetab \mid \mathbb{D}_{k-1})$ from the updated belief distribution and (2) how to compute the sequential LFIRE ratio in~\eqref{eq:seqratio} given the observations $\mathbb{D}_{k-1}$. We explain our solutions to these difficulties in Sections~\ref{sec:upd} to~\ref{sec:seqlfire}.





\subsection{Updating the belief about the model parameters} \label{sec:upd}

For iteration $k=1$ we only require samples from the prior
distribution $p(\thetab)$ in order to compute the MI
in~\eqref{eq:sequtility_mc}. We here assume that sampling from the
prior is possible. For iteration $k=2$, we require samples from
$p(\thetab \mid \mathbb{D}_1)$, for $k=3$ we require samples from
$p(\thetab \mid \mathbb{D}_2)$, etc.
We here describe how to obtain samples from the updated belief $p(\thetab \mid \mathbb{D}_{k})$ after any iteration $k$. For that, let us first define what it means to update the belief about the model parameters. After observing real-word data $\ybf^\ast_{k}$ at optimal design $\dbf^\ast_{k}$, we update the observation data set, i.e.~$\mathbb{D}_{k} = \mathbb{D}_{k-1} \cup \{\dbf_{k}^\ast, \ybf^\ast_{k}\}$. 
For $\dbf = \dbf^\ast_{k}$ and $\ybf=\ybf^\ast_{k}$, the numerator in~\eqref{eq:seqratio} equals $p(\thetab \mid \mathbb{D}_{k})$, leading us to an expression for the updated belief distribution, 
\begin{align}
p(\thetab \mid \mathbb{D}_{k}) &\approx \widehat{r}_{k}(\dbf_{k}^\ast, \ybf_{k}^\ast, \thetab, \mathbb{D}_{k-1}) p(\thetab \mid \mathbb{D}_{k-1}). \label{eq:up_1}
\end{align}
Furthermore, we can approximate the belief distribution $p(\thetab \mid \mathbb{D}_{k})$ after iteration $k$ as a product of $k$ estimated density ratios and the initial prior $p(\thetab)$,
\begin{align}
p(\thetab \mid \mathbb{D}_{k}) 
&\approx \widehat{r}_{k}(\dbf_{k}^\ast, \ybf_{k}^\ast, \thetab, \mathbb{D}_{k-1}) \cdots \widehat{r}_{1}(\dbf_{1}^\ast, \ybf_{1}^\ast, \thetab) p(\thetab). \label{eq:expand}
\end{align}
Each of the density ratios $\widehat{r}_s$ in~\eqref{eq:expand} are evaluated at the observations $\{\dbf_s^\ast, \ybf_s^\ast\}$ of the relevant iteration $s$, but also depend on all previous observations $\mathbb{D}_{s-1}$. We can write this product of density ratios as a weight function $w_k$ and then~\eqref{eq:expand} becomes
\begin{align}
p(\thetab \mid \mathbb{D}_{k}) &\approx w_k(\thetab; \mathbb{D}_{k}) p(\thetab),
\end{align}
where we have defined the weight function $w_k$ to be
\begin{align}
w_k(\thetab; \mathbb{D}_{k}) = \prod_{s=1}^{k} \widehat{r}_{s}(\dbf_{s}^\ast, \ybf_{s}^\ast, \thetab, \mathbb{D}_{s-1}), \label{eq:weights}
\end{align}
%
with $\widehat{r}_{1}(\dbf_{1}^\ast, \ybf_{1}^\ast, \thetab, \mathbb{D}_{0}) = \widehat{r}_{1}(\dbf_{1}^\ast, \ybf_{1}^\ast, \thetab)$ according to~\eqref{eq:ratio} and $w_0(\thetab) = 1 \, \forall \, \thetab$.

We use the weight function in~\eqref{eq:weights} to obtain samples from the updated belief distribution $p(\thetab \mid \mathbb{D}_{k})$. To do so, we first sample $N$ initial prior samples $\thetab^{(i)} \sim p(\thetab)$. After every iteration $k$ we then obtain weights $w_{k}^{(i)} = w_{k}(\thetab^{(i)}; \mathbb{D}_{k})$ corresponding to the initial prior samples, which form a particle set $\{w_{k}^{(i)}, \thetab^{(i)}\}_{i=1}^{i=N}$. We compute these by updating each weight $w_{k-1}^{(i)}$ by the LFIRE ratio evaluated at the observed data according to~\eqref{eq:weights}, in order to yield $w_{k}^{(i)}$. Since we can store the weights $w_{k-1}^{(i)}$ for a particular parameter, we do not need to recompute them. 

To finally obtain updated belief samples, we first normalise the weights: $W_{k}^{(i)} = w_{k}^{(i)} /\, \Sigma_{i=1}^{N} w_{k}^{(i)}$. Then we sample an index from the categorical distribution, i.e.~$I \sim \text{cat}(\{W_{k}^{(i)}\})$, and choose the initial prior sample $\thetab^{(I)}$. Repeating this several times results in a set of parameter samples that follows $p(\thetab \mid \mathbb{D}_{k})$. 
We have summarised this procedure in Algorithm~\ref{algo:postsamples}.
\begin{algorithm}
\caption{Obtaining samples from the updated belief $p(\thetab \mid \mathbb{D}_{k})$}\label{algo:postsamples}
\begin{algorithmic}[1]
\State {After iteration $k$, obtain particle set $\{w_{k}^{(i)}, \thetab^{(i)}\}_{i=1}^{i=N}$}
\State {Normalise the weights: $W_{k}^{(i)} = w_{k}^{(i)} /\, \Sigma_{i=1}^{N} w_{k}^{(i)}$ for $i=1, \dots, N$}
\For {$i=1$ to $i=N$}
  \State Sample from a categorical distribution: $I \sim \text{cat}(\{W_{k}^{(i)}\})$
  \State Choose $\thetab^{(I)}$ as a sample from the updated prior distribution
\EndFor
\end{algorithmic}
\end{algorithm}

We note that approximating~\eqref{eq:sequtility_implicit} directly
with weighted samples from $p(\thetab)$ instead of using
Algorithm~\ref{algo:postsamples} to obtain samples from $p(\thetab
\mid \mathbb{D}_{k-1})$ would theoretically result in a lower variance
Monte-Carlo estimator. However, because we did not observe this in
our simulations and Algorithm~\ref{algo:postsamples} had lower computation
times, we opted to use Algorithm~\ref{algo:postsamples} instead. For
more details see Appendix~\ref{app:B}.

\subsection{Resampling} \label{sec:resampling}

We can see from~\eqref{eq:weights} that a weight $w_k^{(i)}$ is
computed by the product of LFIRE ratios given all previous
observations. Less significant parameter samples, i.e.~ones with a low
density ratio, thus have weights that quickly decay to zero. This
means that after several iterations we may be left with only a few
weights $w_k^{(i)}$ that are effectively non-zero. Eventually, only
few different parameter samples $\thetab^{(i)}$ of the current
particle set are chosen in the sampling scheme
in Algorithm~\ref{algo:postsamples}, which increases the
Monte-Carlo error of the sequential utility
in~\eqref{eq:sequtility_mc} and of the marginal samples in the
sequential LFIRE procedure (see Section~\ref{sec:seqlfire}).

We can quantify how many effective samples we have via the \emph{effective sample size} $\eta$~\citep{Kish1965},
\begin{equation}
\eta = \frac{\left(\sum_{i=1}^N w_k^{(i)}\right)^2}{\sum_{i=1}^N \left(w_k^{(i)}\right)^2}. \label{eq:ess}
\end{equation}
If $\eta$ is small, i.e.~$\eta \ll N$, then they do not cover much
relevant parameter space and our Monte-Carlo approximations may become
poor. Thus, if the effective sample size becomes smaller than a
minimum sample size $\eta_{\text{min}}$ we need to resample our set of
parameter samples; this allows us to have a set of new parameter
samples that well-represent the current belief distribution. From a
practical point of view, throughout this work we shall use the typical
value of $\eta_{\text{min}} = N/2$~\citep[e.g.][]{Chen2003,
  Doucet2009}.

We start the resampling procedure by transforming the parameter space such that all parameter samples $\thetab^{(i)}$ have values between $0$ and $1$, i.e.~$\thetab \rightarrow \thetab^\prime$. This ensures that different parameter dimensions have similar scales, thereby increasing the robustness of the following steps.\footnote{See Appendix~\ref{app:C} for a more detailed explanation.} If the parameter space has boundary conditions $\mathcal{B}$, through a bounded prior distribution for instance, then we transform these in the same way as the parameter space, $\mathcal{B} \rightarrow \mathcal{B}^\prime$ (see Appendix~\ref{app:C}). In this transformed space, we model the belief distribution $p(\boldsymbol{\theta}^\prime \mid \mathbb{D}_{k})$ after the current iteration $k$ as a truncated Mixture of Gaussians (MoG), i.e.~
\begin{equation} \label{eq:mog}
p(\boldsymbol{\theta}^\prime \mid \mathbb{D}_{k}) \approx \sum_{i=1}^N W_k^{(i)} \mathcal{N}(\thetab^\prime; \thetab^{\prime(i)}, \mathbb{I}\sigma^2) \mathds{1}_{\mathcal{B}^{\prime}}(\thetab^{\prime}),
\end{equation}
%
where $\mathbb{I}$ is the identity matrix, $\sigma^2$ is a variance
parameter and $W_k^{(i)}$ are the normalised weights. The indicator
function $\mathds{1}_{\mathcal{B}^{\prime}}(\thetab^{\prime})$ is $1$
if $\thetab^{\prime}$ satisfies the boundary conditions
$\mathcal{B}^{\prime}$ and $0$ otherwise. Note that we have one
Gaussian for every parameter sample of the current particle set; each
Gaussian is centred at that parameter sample $\thetab^{\prime(i)}$
and has the same standard deviation $\sigma$. The parameter $\sigma$
is typically small which means that~\eqref{eq:mog} should be
understood more in terms of smoothing than Gaussian mixture
modelling.

We compute the Gaussian standard deviation by first using a k-dimensional (KD) tree~\citep{Bentley1975} to find the nearest neighbour $\text{NN}(\thetab^{\prime(i)})$ of each parameter sample. Let $\delta$ be the median of all the distances of a sample to its nearest neighbour, i.e.~$\delta = \text{median}\left( \lvert \thetab^{\prime(i)} - \text{NN}(\thetab^{\prime(i)}) \rvert \right)$. We then compute the standard deviation $\sigma$ as a function $g$ of $\delta$, i.e.~
\begin{equation}
\sigma = g(\delta), \label{eq:sigma}
\end{equation}
where we choose $g$ to be the square-root function in order to increase robustness to possibly large median distances.\footnote{The log function would work similarly well for this reason.}

In order to get a new sample from the updated belief distribution, we
sample an index from a categorical distribution, i.e.~$I \sim
\text{cat}(\{W_k^{(i)}\})$, and obtain a parameter sample from the
corresponding Gaussian $\mathcal{N}(\thetab^\prime;
\thetab^{\prime(I)}, \mathbb{I} \sigma^2)$. We accept this
parameter sample if it satisfies the transformed boundary conditions
$\mathcal{B}^\prime$ and reject it otherwise. Doing this a number of
times yields a set of new parameter samples.
We then set the weight of each of the new, resampled parameter samples as proportional to one and transform the samples back to the original parameter space, i.e.~$\thetab^\prime \rightarrow \thetab$. This procedure of resampling parameters is summarised in Algorithm~\ref{algo:resampling}.

Unlike the resampling step that uses MCMC in previous approaches to SMC sequential design~\citep{Hainy2016b}, our approach does not exactly preserve the distribution of the particles. However, crucially, it does not require re-processing all data collected to date, accelerating computation.
Conceptually, our resampling method can also be understood as fitting some model to weighted samples from the prior distribution. This can be viewed as a type of kernel density estimate (KDE) formed from weighted samples. 
Other density estimators that allow for sampling, even fully-parametric ones, could be used as well. 

\begin{algorithm}
\caption{Resampling via a Mixture of Gaussian model\label{algo:resampling}}
\begin{algorithmic}[1]
\State {After iteration $k$, obtain particle set $\{w_k^{(i)}, \thetab^{(i)}\}_{i=1}^{i=N}$}
\State {Transform the parameters to be in the unit hyper-cube, $\thetab \rightarrow \thetab^\prime$}
\State {Transform the boundary conditions in the same way, $\mathcal{B} \rightarrow \mathcal{B}^\prime$}
\State {Find the nearest neighbour of each parameter sample}
\State {Compute the standard deviation $\sigma$ for the MoG model, according to~\eqref{eq:sigma}}
\State {Normalise the weights: $W_k^{(i)} = w_k^{(i)} /\, \Sigma_{j=1}^{N} w_k^{(j)}$ for $i=1, \dots, N$}
\For {$i=1$ to $i=N$}
  \State Sample from a categorical distribution: $I \sim \text{cat}(\{W_k^{(i)}\})$
  \While {not accepted}
	  \State Sample $\thetab^{\prime(i)}_{new} \sim \mathcal{N}(\thetab^\prime; \thetab^{\prime(I)}, \mathbb{I} \sigma^2)$
	  \If {$\thetab^{\prime(i)}_{new}$ satisfies $\mathcal{B}^\prime$}
	  	\State {Accept}
	  \Else
	  	\State {Reject}
	  \EndIf
  \EndWhile
\EndFor
\State {Reset the weights to $w_k^{(i)} = 1 \, \forall \, i$}
\State {Transform the parameters back to the original parameter space, $\thetab^\prime_{new} \rightarrow \thetab_{new}$}
\State {Return $\{w_k^{(i)}, \thetab^{(i)}_{new}\}_{i=1}^{i=N}$} 
\end{algorithmic}
\end{algorithm}

\subsection{Sequential LFIRE} \label{sec:seqlfire}
As can be seen from~\eqref{eq:seqratio}, the sequential LFIRE ratios depend on previous observations. This particular dependency requires us to revise the original LFIRE method of~\citet{Thomas2016} slightly. To compute the ratio $\widehat{r}_{k}(\dbf, \ybf, \thetab, \mathbb{D}_{k-1})$ we need to sample data from the likelihood $p(\ybf \mid \thetab, \dbf)$ and from the marginal $p(\ybf \mid \dbf, \mathbb{D}_{k-1})$. We again assume that observing data does not affect the data generating process, i.e.~sampling from the likelihood remains unchanged. The marginal distribution does change upon observing data, i.e.~at iteration $k$ we have
\begin{align}
p(\ybf \mid \dbf, \mathbb{D}_{k-1}) &= \int p(\ybf, \thetab \mid \dbf, \mathbb{D}_{k-1}) \mathrm{d}\thetab \\
&= \int p(\ybf \mid \thetab, \dbf) p(\thetab \mid \mathbb{D}_{k-1}) \mathrm{d}\thetab. \label{eq:marg}
\end{align}
This implies that in order to obtain samples from the marginal we first have to sample from the belief distribution $p(\thetab \mid \mathbb{D}_{k-1})$ according to Algorithm~\ref{algo:postsamples}. These parameter samples from the updated belief distribution are then plugged into the data generating distribution to finally obtain samples from the marginal. The rest of the LFIRE procedure remains unchanged~\citep[see][for more details]{Thomas2016}.

\subsection{Optimisation} \label{sec:bayesopt}

In all sections hitherto we have explained how to compute the sequential mutual information utility $\widehat{U}_k(\dbf)$ at iteration $k$. We have, however, not addressed the issue of optimising the utility with respect to the designs $\dbf$ in order to find the optimal design $\dbf_k^\ast$. While traditionally the utility has been optimised via grid search or a sampling-based approach by~\citet{Muller99}, there have been a few recent approaches using evolutionary algorithms~\citep{Price2018} or Gaussian Processes (GP)~\citep{Overstall2017}. The latter approaches were generally found to outperform grid search in terms of efficiency. We here choose to optimise the sequential utility using Bayesian Optimisation (BO)~\citep{Shahriari2016}, as was done by~\citet{Kleinegesse2019}, due to its flexibility and efficiency. In addition, BO smoothes out the Monte-Carlo error of our utility approximations, and may therefore help in locating the optimal design $\dbf^\ast_k$ as well.

BO is a popular optimisation scheme for functions that are expensive to evaluate and that potentially have unknown gradients. The general idea is to use a probabilistic surrogate model of the utility and then use a cheaper acquisition function to decide where to evaluate the utility next. We use a GP for the surrogate model with a Mat\'ern-5/2 Kernel~\citep{Shahriari2016} and Expected Improvement~\citep{Mockus1978} for the acquisition function. These are standard choices in the BO literature, for more detail see~\citet{Shahriari2016}.

\pagebreak
We summarise the previous sections by describing our framework of estimating and optimising the sequential mutual information utility in Algorithm~\ref{algo:seqdesign}. 
\begin{algorithm}[H]
\caption{Sequential Bayesian Exp. Design via LFIRE using BO\label{algo:seqdesign}}
\begin{algorithmic}[1]
\State {Let $\mathbb{D}_{0} = \varnothing$}
\State {Sample initial parameters from prior: $\boldsymbol{\theta}^{(i)} \sim p(\boldsymbol{\theta})$ for $i=1, \dots, N$}
\State {Initialise weights: $w_0^{(i)} = 1$ for $i=1, \dots, N$}
\For {$k=1$ to $k=K$}
  \State {Calculate the effective sampling size $\eta$ using~\eqref{eq:ess}}
  \If {$k=1$}
    \State {Use all initial prior samples $\{\thetab^{(i)}\}_{i=1}^N$}
  \ElsIf {{$\eta < \eta_{\text{min}}$}}
  	\State {Obtain new samples $\{\thetab^{(i)}_{new}\}_{i=1}^N$ by resampling according to Algorithm~\ref{algo:resampling}}
    \State {Set the weights for the iteration to one, i.e. $w_{k}^{(i)} = 1$ for $i=1, \dots, N$}
    \State {Use all new parameter samples $\{\thetab^{(i)}\}_{i=1}^N \leftarrow \{\thetab^{(i)}_{new}\}_{i=1}^N$}
  \Else
  	\State {Obtain updated belief samples $\{\thetab^{(i)}\}_{i=1}^N$ by applying Algorithm~\ref{algo:postsamples}}
  	\State {Use all updated belief samples $\{\thetab^{(i)}\}_{i=1}^N$}
  \EndIf
  \State {Use BO to determine the maximiser $\dbf^\ast_{k}$ of the sequential utility $\widehat{U}_k(\dbf)$ in~\eqref{eq:sequtility_mc}}
  \State {Perform an experiment at $\dbf_k^\ast$ to observe some real data $\ybf^\ast_{k}$}
  \State {Update the belief distribution by updating the data set: $\mathbb{D}_{k} = \mathbb{D}_{k-1} \cup \{\dbf_k^\ast, \ybf^\ast_{k}\}$}
  \State {For all parameter samples $\thetab^{(i)}$, compute new weights $w_{k}^{(i)}$ according to~\eqref{eq:weights}}
\EndFor \relax
\end{algorithmic}
\end{algorithm}


\section{Experiments} \label{sec:experiments}

In this section we test the framework outlined in Algorithm~\ref{algo:seqdesign} on a number of implicit models from the literature. We first consider an oscillatory toy model with a multi-modal posterior distribution. We then consider the Death Model~\citep{Cook2008} and the SIR Model~\citep{Allen2008} from epidemiology, as well as a model of the spread of cells~\citep{Vo2015}. 
We evaluate these models with our framework that approximates the sequential MI utility with density ratio estimation, i.e. LFIRE. 

\subsection{Oscillation Toy Model}

This toy model describes noisy measurements of a sinusoidal, stationary waveform $\sin(\omega t)$, where the design variable is the measurement time $t$ and the experimental aim is to optimally estimate the waveform's frequency $\omega$. The generative model is given by
\begin{align}
p(y \mid \omega, t) = \mathcal{N}(y; \sin(\omega t), \sigma_{\text{noise}}^2), \label{eq:sinean}
\end{align}
where we set the measurement noise to $\sigma_{\text{noise}} = 0.1$ throughout and assume that the true model parameter takes a value of $\omega_{\text{true}}=0.5$. As a prior we use a uniform distribution $p(\omega) = \mathcal{U}(\omega; 0, \pi)$. We can obtain analytic posterior densities by using the likelihood in~\eqref{eq:sinean} and Bayes' rule, while we can obtain corresponding posterior samples by using Markov chain Monte-Carlo (MCMC) methods. 

We start the sequential BED procedure
for the oscillation model by sampling $1{,}000$ parameter
samples $\omega^{(i)}$ from the prior and for each of these we then simulate data
$y^{(i)} \sim \mathcal{N}(y; \sin(\omega^{(i)} t),
\sigma_{\text{noise}}^2)$ at a particular measurement time $t \in [0,
  2\pi]$. For the summary statistics in~\eqref{eq:lfire_ratio} we use
subsequent powers of the simulated data, i.e. $\bm{\psi}(y^{(i)}) =
[y^{(i)}, (y^{(i)})^2, (y^{(i)})^3]^\top$, in order to allow for a
sufficiently flexible, non-linear decision boundary in the LFIRE
algorithm. We use these prior samples and the corresponding simulated
data to compute 1{,}000 LFIRE ratios and then estimate the MI utility
$\widehat{U}_1(t)$ with a sample average as
in~\eqref{eq:sequtility_mc}. With the help of BO, we decide at which
measurement time $t$ to evaluate the utility next and then repeat
until we have maximised the utility, following
Algorithm~\ref{algo:seqdesign}.

We repeat the optimisation procedure above for the BD-Opt utility
in~\eqref{eq:bdopt} and compare it to the MI
utility. \citet{Hainy2016b} used this utility in sequential design
targeted at implicit models, although they only tested their method on
a toy model with known likelihood. The advantages of MI over BD-Opt
for models with multi-modal posteriors are widely known in the
explicit setting~\citep{Ryan2016}. It is nonetheless useful to verify
that these advantages continue to hold when approximating the MI and the
posterior with LFIRE.

%
\begin{figure}[!t]
\includegraphics[width=\linewidth]{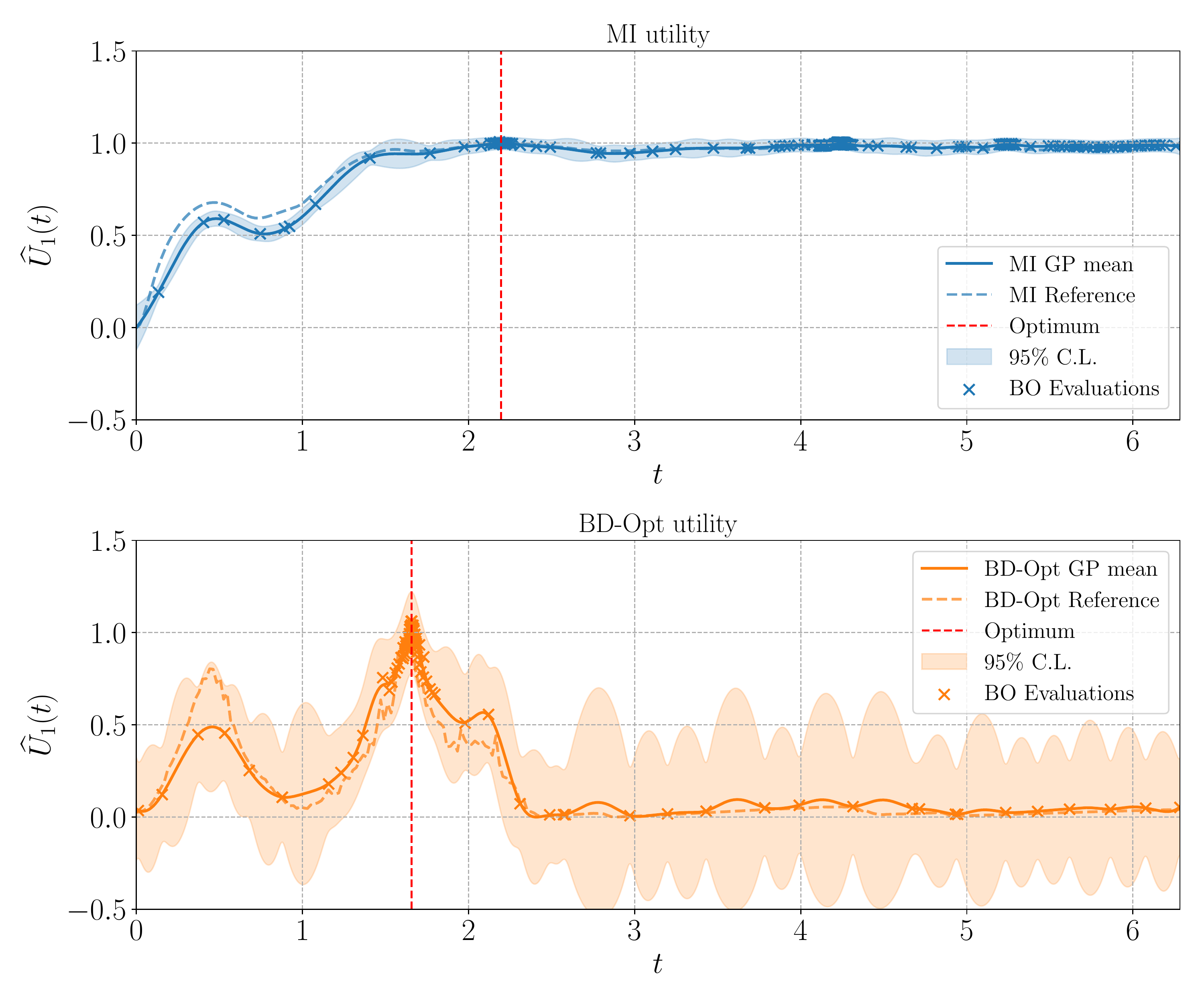}
\caption[]{Comparison of MI (top) and BD-Opt (bottom) utilities for the first iteration of the oscillatory toy model, including analytic references.}
\label{fig:sine_utils_iter1}
\end{figure}

We show the MI utility and the BD-Opt utility used
by~\citet{Hainy2016b}, as well as their analytic counterparts, for the
first iteration in Figure~\ref{fig:sine_utils_iter1}. Shown are the
posterior predictive means of the GP surrogate models, the
corresponding variances and the evaluations of the utilities during
the BO procedure.
Due to the chosen prior and the periodic nature of the oscillation model, higher design
times result in posterior distributions with more modes.
Multi-modality can lead to an increase of the variance. BD-Opt thus
assigns little to no worth in doing experiments at late measurement
times.
In contrast, the MI utility has a high value at late design times
when the posterior distributions tend to have multiple modes.
The corresponding optimal designs are $t_1^\ast = 2.196$ and $t_1^\ast = 1.656$ for the MI utility and the BD-Opt utility, respectively. Furthermore, the behaviour of both utilities generally well matches the analytic references computed using the closed-form expression of the data-generating distribution.

\begin{figure}[!t]
\includegraphics[width=\linewidth]{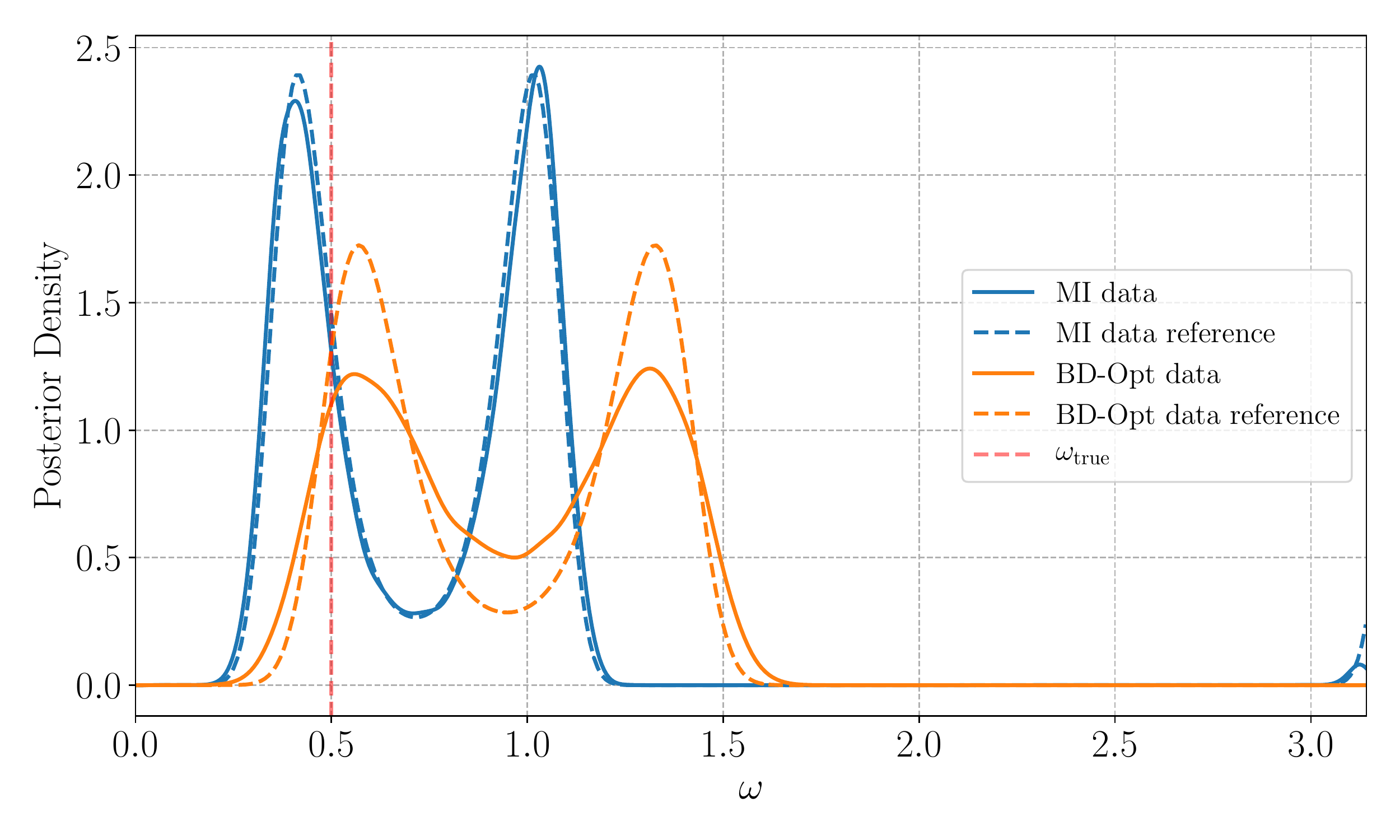}
\caption[]{Comparison of the posterior distributions obtained when using the data from the MI (blue) and BD-Opt (orange) utilities for the first iteration of the oscillatory toy model (solid curves), including analytic references (dashed curves).}
\label{fig:sine_postan_iter1}
\end{figure}

After determining the optimal measurement time $t_1^\ast$, we perform
the actual experiment. Here, the real-world experiment is simulated by
taking a measurement of the true data generating process with
$\omega_{\text{true}}=0.5$ at $t_1^\ast$ where we obtained
$y_1^\ast=0.790$ and $y_1^\ast=0.810$ for the case of the MI and
BD-Opt utility, respectively. We show the corresponding estimates of
the posterior distributions in Figure~\ref{fig:sine_postan_iter1}. In
our approach, we compute particle weights for each of the 1{,}000
prior samples and then obtain the posterior, or updated belief,
samples according to Algorithm~\ref{algo:postsamples}. Importantly,
the BD-Opt utility uses a particle approach as well, which means that
we also need to use Algorithm~\ref{algo:postsamples} to obtain updated
belief samples; we direct the reader to~\citet{Hainy2016b} for more
information on how the required particle weights are computed. For
visualisation purposes, we then compute a Gaussian Kernel Density
Estimate (KDE) from these posterior samples to obtain the posterior
densities shown in Figure~\ref{fig:sine_postan_iter1}. We also show
the analytic posteriors which are computed using the closed-form
expression of the data-generating distribution. 
We find that the posterior distributions have two modes, which is a
result of the periodic behaviour of the model.
We note that one mode has support for the true model parameter.

After obtaining the relevant data $\mathbb{D}_1 = \{\dbf_1^\ast, \ybf_1^\ast \} = \{t_1^\ast, y_1^\ast \}$ for the first iteration $k=1$, we compute the new particle weights $w_1^{(i)}$ via~\eqref{eq:weights} for MI and via ABC likelihoods for BD-Opt~\citep[see][]{Hainy2016b}, which are then used in subsequent iterations of the sequential BED procedure. 
Following Algorithm~\ref{algo:seqdesign} we continue this procedure in a similar manner until iteration $k=4$, although technically this could be continued until the experiment's budget is exhausted. 

\begin{figure}[!t]
\includegraphics[width=\linewidth]{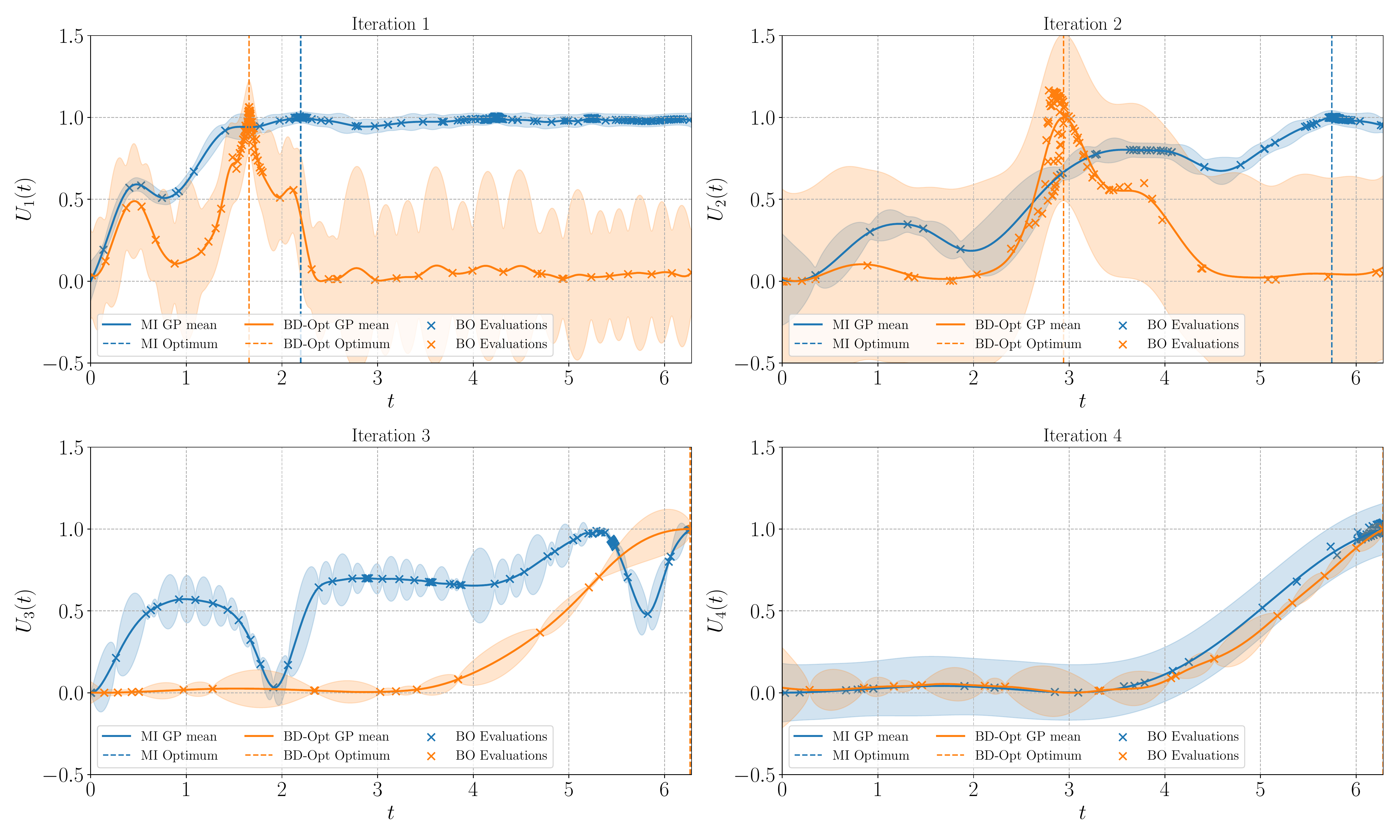}
\caption[]{Comparison of the MI and BD-Opt utilities at different iterations for the oscillatory toy model. Shown are the GP means and variances, the BO evaluations and the optima of the GP means. After sufficient iterations, 4 in this case, the difference between the two utilities becomes negligible.}
\label{fig:sine_utils}
\end{figure}

We show the GP models of the MI and BD-Opt utility for all
four iterations in Figure~\ref{fig:sine_utils}.
Both utilities change vastly between iterations. As compared to iteration 1, the MI utility has more local optima in iteration 2, although it is still overall increasing and then peaking at $t\approx6$. A pronounced local minimum occurs around $t=2.196$, the optimal design of the first iteration; this is intuitive, because performing an experiment at the same experimental design may not yield much additional information at this stage due to the relatively small measurement noise.
For the same reason, the BD-Opt utility has a local minimum around $t=1.656$, the optimal design of the first iteration for BD-Opt. Due to large fluctuations in the estimated BD-Opt utility around the global maximum, the GP mean does not go through all nearby evaluations and has a larger variance throughout for iteration 2. 

In iteration 3, the MI utility has two local minima that occur at the locations of the two previous optimal designs because, like previously, performing an experiment at the same measurement locations may not be effective. BD-Opt on the other hand steadily increases and then peaks at the upper boundary of the design domain. This occurs because, for BD-Opt, the updated belief distribution of the parameter is uni-modal after iteration $2$ and becomes more narrow with increasing design times; similar reasoning follows for BD-Opt in iteration $4$. We observe the same behaviour for MI in iteration $4$, as the updated belief distribution used to compute the MI utility becomes uni-modal after iteration $3$.
We had to perform resampling during iterations $2-4$, as the effective
sample size of the weights went below 50\% for both the MI and the
BD-Opt utility.\footnote{Note that for BD-Opt we use the resampling
  procedure provided in~\citet{Hainy2016b}} Figure
\ref{fig:sine_utils} shows that in iteration 1 to 3, mutual
information assigns worth to several areas in the design domain that
Bayesian D-Optimality does not deem important. After enough data is
collected and the posterior is unimodal, however, the difference
between these two utilities becomes negligible and they result in the
same optimal design.

\begin{figure}[!t]
\includegraphics[width=\linewidth]{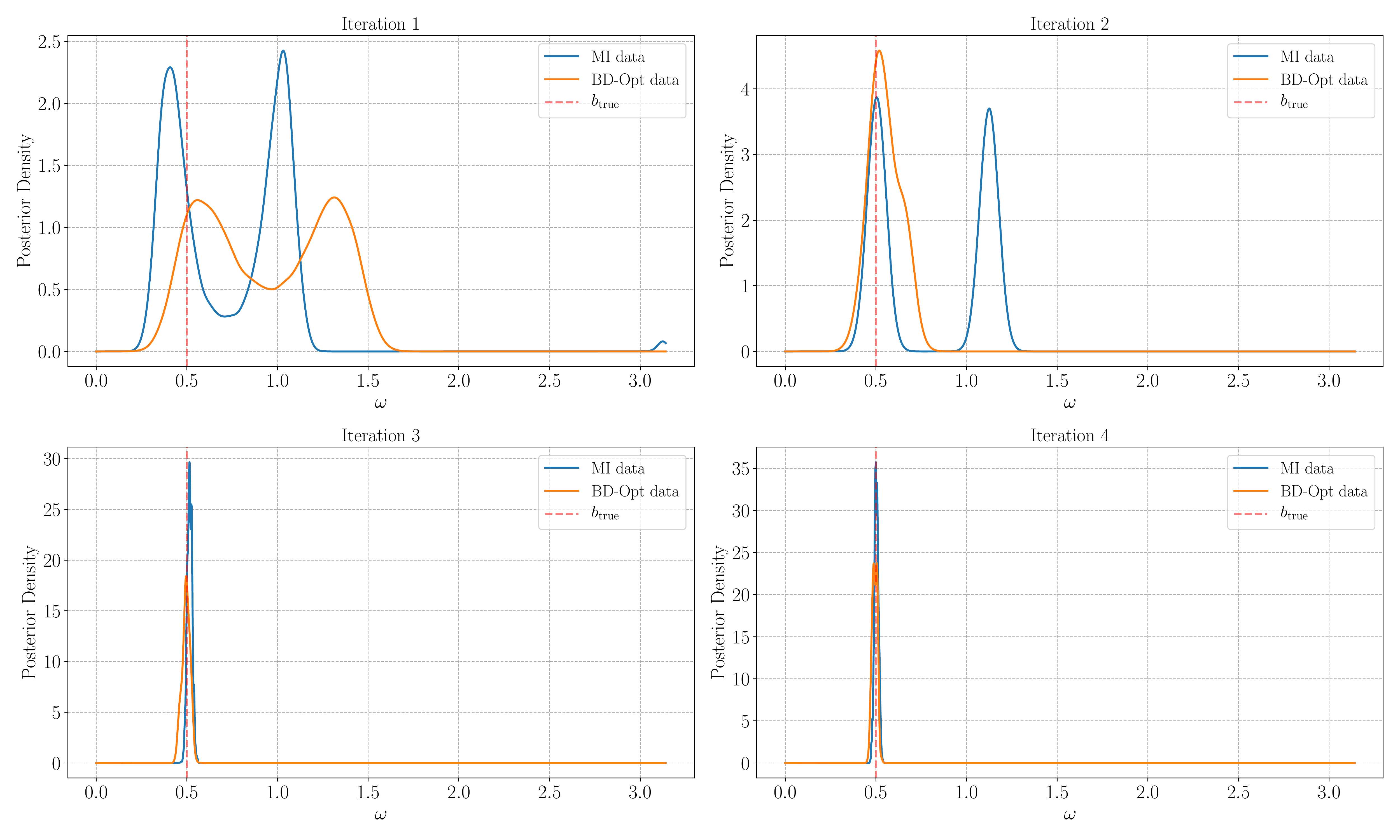}
\caption[]{Comparison of the posterior distributions obtained when using the data from the MI and BD-Opt utilities for four iterations of the oscillatory toy model.}
\label{fig:sine_posts}
\end{figure}

For visualisation purposes, we put a KDE over updated belief samples
after each iteration, obtained by means of
Algorithm~\ref{algo:postsamples}, to plot posterior densities. This is
shown in Figure~\ref{fig:sine_posts} for the sequential MI and BD-Opt
utilities. After iteration 2, only mutual information results in a
multi-modal belief distribution. From iteration 3 onwards, both
distributions are unimodal and similarly concentrated around the true
model parameter of $\omega_{\text{true}}=0.5$. After 4 iterations, the
mean parameter estimate using the data from the MI utility is
$\widehat{\omega}=0.503$ with a $95\%$ credibility interval of
$[0.481, 0.527]$. Using the data from the BD-Opt utility the mean
parameter is $\widehat{\omega}=0.494$ with a $95\%$ credibility
interval of $[0.468,0.516]$. The $95\%$ credibility intervals where
computed using a Gaussian KDE of the parameter samples and the highest
posterior density interval (HPDI) method.

Overall, in the context of the oscillation model, the mutual
information and Bayesian D-Optimality utilities yield significantly
different optimal experimental designs. As opposed to MI, BD-Opt leads
to optimal experimental designs that are biased to exclude multiple
explanations for the inferred parameters. When enough real-world
observations are made, the updated belief distributions are no longer
multi-modal but collapse to unimodal distributions, at which point the
utilities become similar. Additionally, for the BD-Opt utility we
noticed certain numerical instabilities that resulted from taking the
mean of several posterior precisions. We rectified this by taking the
median of several posterior precisions, instead of taking the mean as
in~\citet{Hainy2016b}.

\subsection{Death Model}

The Death Model describes the stochastic decline of a population of $N$ individuals due to some infection. Individuals change from the susceptible state $S$ to the infected state $I$ at an infection rate $b$, which is the model parameter we are trying to estimate. Each susceptible individual can get infected with a probability of $p_{\mathrm{inf}}(t) = 1 - \exp(-bt)$~\citep{Cook2008} at a particular time $t$.
The aim of the Death Model is to decide at which measurement times $\tau$ to observe the infected population $I(\tau)$ in order to optimally estimate the true infection rate $b$.\footnote{See Appendix~\ref{app:D} for a time series plot of the Death Model.} Here we assume that for each iteration of the sequential BED scheme we only have access to a new, independent stochastic process. This means that, for instance, in iteration $k=2$ we could have design times before the optimal design time $\tau^\ast_1$ of the first iteration. 

The total number of individuals $\Delta I(t)$ moving from state $S$ to state $I$ at time $t$ is given by a sample from a Binomial distribution,
\begin{align}
\Delta I (t) \sim \mathrm{Bin}(\Delta I (t); N - I(t), p_{\mathrm{inf}}(\Delta t)), \label{eq:death1}
\end{align}
where  $\Delta t$ is the step size, set to $0.01$ in this work, and $I(t = 0) = 0$. By discretising this time series, the number of infected at time $t + \Delta t$ is given by $I(t+\Delta t) = I(t) + \Delta I (t)$. The likelihood for this model is analytically tractable~\citep[see][]{Cook2008, Kleinegesse2019}, and thus can serve as a means to validate our framework. As a prior distribution we use a truncated Normal distribution, centred at $1$ with a standard deviation of $1$, while the summary statistics used to compute~\eqref{eq:lfire_ratio} are subsequent powers of the number of infected, i.e.~$\bm{\psi}(I(\tau))=[I(\tau), I(\tau)^2, I(\tau)^3]^\top$. To generate real-world observations, we use a true parameter value of $b_{\text{true}}=1.5$.

\begin{figure}[!t]
\includegraphics[width=\linewidth]{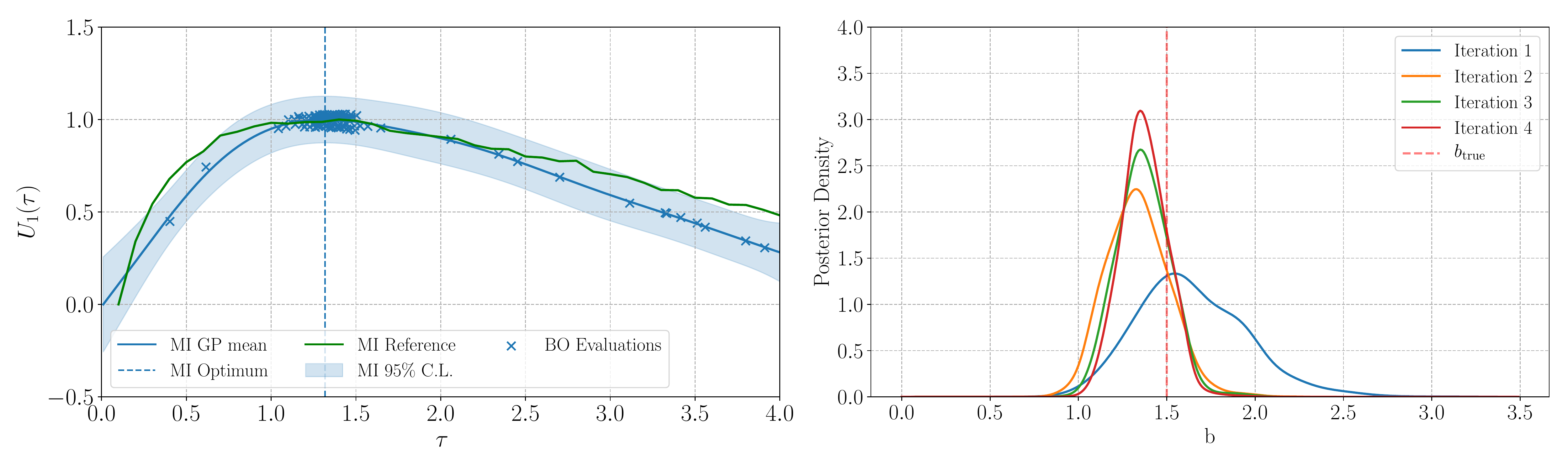}
\caption[]{Left: MI utility for the first iteration of the Death model, including a reference MI computed numerically; shown are the GP mean and variance, BO evaluations and optimum of the GP mean. Right: Updated belief distributions after different iterations of the proposed sequential Bayesian experimental design approach for the Death model.}
\label{fig:death_utils}
\end{figure}

We show the first iteration of the sequential mutual information utility in the left plot of Figure~\ref{fig:death_utils}, as well as a reference MI value obtained using the tractable likelihood.\footnote{See Appendix~\ref{app:E} for a derivation of the reference MI.} The MI peaks in the region around $\tau \approx 1$ and stays low at early and late measurement times. The average posterior for early and late $\tau$ is wider than the one for $\tau \approx 1$,\footnote{We show posterior plots for different measurement times in Appendix~\ref{app:F}.} which results in a lower MI at the boundary regions. This is because at early and late measurement times the observed number of infected $I(\tau)$ is the same for a wide range of infection rates $b$, i.e.~either $0$ or $50$ (the extreme values of $I(\tau)$). At $\tau \approx 1$ most values of $b$ yield observations of $I(\tau)$ that are between the extreme values $0$ and $50$, allowing us to infer the relationship between $b$ and $I(\tau)$ more effectively. For later iterations, the MI generally had the same form and did not change much, with optimal measurement times that were all roughly around $1$ (see Appendix~\ref{app:F} for a plot with all iterations). This reduces the uncertainty in $b$ which, in this case, outweighs the advantages of making an observation at different measurement times such as near the boundaries.


%

We show a KDE of the updated belief samples after each iteration,
obtained by means of Algorithm~\ref{algo:postsamples}, in the right
plot of Figure~\ref{fig:death_utils}. Even though the updated belief
distribution after the first iteration has an expected value that is
close to the true parameter, the corresponding credibility interval is
wide. The belief distributions in the following iterations become more
narrow, which is a result of having more data to estimate the
model parameter. 
After four iterations, the posterior mean of $b$ equals $\widehat{b} =
1.376$ with a $95\%$ credibility interval of $[1.128, 1.621]$
containing $b_{\text{true}}=1.5$. The credibility intervals were
computed using a Gaussian KDE over posterior samples and the HPDI
method.




\subsection{SIR Model}

The SIR Model~\citep{Allen2008} is an extension of the Death model and, in addition to the number of susceptibles $S(t)$ and infected $I(t)$, includes one more state population, the number of individuals $R(t)$ that have recovered from the infection and cannot be infected again. Similar to the Death model, the design variable is the measurement time $\tau$ at which to observe the state populations. For this model however, we are trying to estimate two model parameters, the rate of infection $\beta$ and the rate of recovery $\gamma$. Similar to the Death model, we assume that for each iteration of the sequential BED scheme we only have access to a new, independent stochastic process.

At a particular time $t$ of the time-series of state populations, let the number of individuals that get infected during an interval $\Delta t$, i.e.~change from state $S(t)$ to state $I(t)$, be $\Delta I(t)$. Similarly, let the number of infected that change to the recovered state be $\Delta R(t)$. We compute these two state population changes by sampling from Binomial distributions, 
\begin{align} \label{eq:statechanges}
\Delta I(t) &\sim \text{Bin}(S(t), p_{\text{inf}}(t)) \\
\Delta R(t) &\sim \text{Bin}(I(t), p_{\text{rec}}(t)),
\end{align}
where the probability $p_{\text{inf}}(t)$ of a susceptible getting infected is defined as $p_{\text{inf}}(t) = \beta I(t) / N$, where $\beta \in [0, 1]$ and $N$ is the total (constant) number of individuals. The probability $p_{\text{rec}}(t)$ of an infected individual recovering from the disease is defined as $p_{\text{rec}}(t)=\gamma$, where $\gamma \in [0, 1]$. These state population changes define the unobserved time-series of the state populations $S$, $I$ and $R$ according to
\begin{align} \label{eq:popchanges}
S(t + \Delta t) &= S(t) - \Delta I(t) \\
I(t + \Delta t) &= I(t) + \Delta I(t) - \Delta R(t) \\
R(t + \Delta t) &= R(t) + \Delta R(t) 
\end{align}
We use initial conditions of $S(t=0) = N-1$, $I(t=0) = 1$ and $R(t=0) = 0$, where we set $N$ to 50 and use a time-step of $\Delta t = 0.01$ throughout. The actual time at which we do observations is again given by $\tau$, such that the observed data is a single value for each state population, i.e.~$S(\tau)$, $I(\tau)$ and $R(\tau)$. We use an uninformative, uniform prior $\mathcal{U}(0,0.5)$ for both model parameters $\beta$ and $\gamma$ to draw initial prior samples. For the summary statistics used to compute~\eqref{eq:lfire_ratio} during the LFIRE algorithm we use subsequent powers, up to 3, of $I(\tau)$ and $R(\tau)$, including their products.\footnote{i.e.~$\psi(\ybf) = [I(\tau), I(\tau)^2, I(\tau)^3, R(\tau), R(\tau)^2, R(\tau)^3, I(\tau)R(\tau), I(\tau)^2R(\tau), I(\tau)R(\tau)^2]^\top$.}
\begin{figure}[!t]
\includegraphics[width=\linewidth]{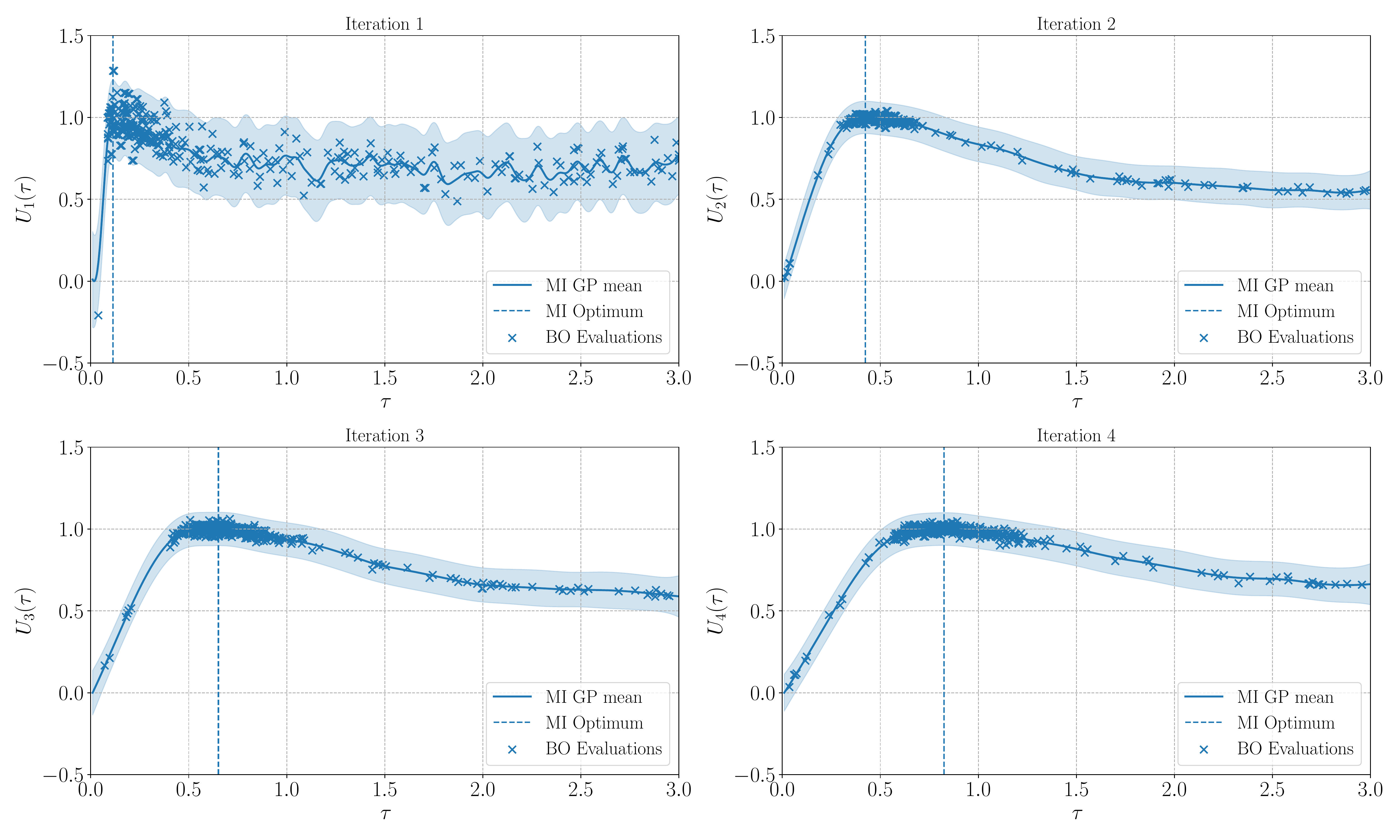}
\caption[]{MI utilities at different iterations for the SIR model. Shown are the GP means and variances, BO evaluations and optima of the GP means.}
\label{fig:sir_utils}
\end{figure}

The first four sequential MI utilities of the sequential BED scheme for the SIR model are shown in Figure~\ref{fig:sir_utils}. 
The SIR model utilities appear similar to those of the Death Model, with the main difference being that the global optima are shifted more towards lower measurement times around $\tau \approx 0.5$, increasing subtlety with every iteration. Similar to the Death model, early and late measurement times result in posterior distributions that are, on average, wider than those for $\tau \approx 0.5$. This is because at early and late $\tau$ much of the data is the same for a wide range of model parameters $\thetab$. At early $\tau$ we mostly observe~$S(\tau)=49$, $I(\tau) = 1$ and $R(\tau)=0$, i.e.~the initial conditions. At late measurement times there are no infected anymore, i.e.~$I(\tau)=0$, and we observe a fixed $S(\tau)$ and $R(\tau)$ that depend on the model parameters (see Appendix~\ref{app:D} for a typical time-series plot). Because the final values of $S(\tau)$ and $R(\tau)$ depend on the model parameters, late measurement times result in posteriors that are slightly more narrow than those for early measurement times. This is reflected in Figure~\ref{fig:sir_utils}, where the MI is higher at late $\tau$ than at early $\tau$. At $\tau \approx 0.5$ we often have numbers of infected $I(\tau)$ that are non-zero, allowing us to infer the relationship between model parameters and data more effectively. This means that the resulting posterior for these measurement times is more narrow than elsewhere, where $I(\tau)$ is close to zero, which leads to the global MI maxima that we see in Figure~\ref{fig:sir_utils}.

Similar to the Death model, the form of the utilities for the SIR
model does not change much between iterations. The utility for the
first iteration appears noisier than the other ones because the
parameter samples during that iteration stem from a uniform prior,
which is highly uninformative and increases the Monte-Carlo error of
the sample average in~\eqref{eq:sequtility_mc}. The other utilities do
not see this issue as the parameter samples from the updated belief
distributions are less spaced out than for the first
iteration. Resampling was performed according to
Algorithm~\ref{algo:resampling} during iterations $2-4$, as the
effective sample size always went below 50\%.
%

%
\begin{figure}[!t]
\includegraphics[width=\linewidth]{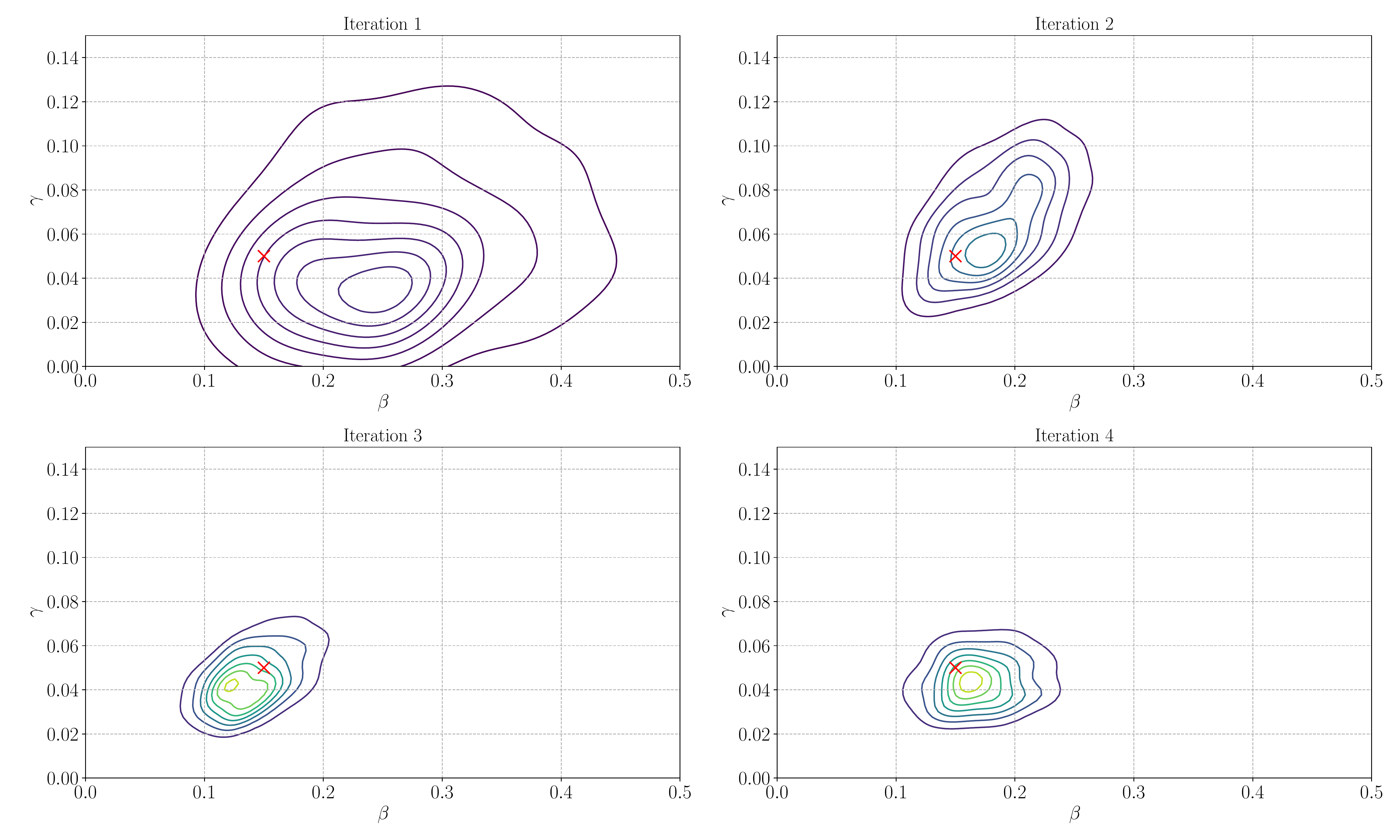}
\caption[]{Comparison of the posterior densities at different iterations for the SIR model. The true model parameters are shown with a red cross.}
\label{fig:sir_posts}
\end{figure}

The posterior densities after every iteration are shown in
Figure~\ref{fig:sir_posts} in form of KDEs computed from the posterior
samples obtained according to Algorithm~\ref{algo:postsamples}.
We see that the beliefs about the model parameters become more precise
after every iteration, which can be attributed due to having more
data. This visualises that data acquired around measurement time $\tau
\approx 0.5$ are providing useful information about the model
parameters.
After four iterations, the mean estimate of the infection rate $\beta$
is $\widehat{\beta} = 0.171$ with a $95\%$ credibility interval of
$[0.112, 0.233]$. The corresponding mean estimate of the recovery rate
$\gamma$ is $\widehat{\gamma}=0.045$ with a $95\%$ credibility
interval of $[0.024, 0.068]$. Similar to before, the credibility
intervals were computed using a Gaussian KDE of the posterior samples
and the HPDI method. The true parameters used to generate real-world
observations were $\beta_{\text{true}} = 0.15$ and
$\gamma_{\text{true}} = 0.05$, which are both contained in the
credibility intervals.

The SIR model example illustrates that we can effectively use the MI utility, computed and optimised via Algorithm~\ref{algo:seqdesign}, to perform sequential BED for an implicit model, where the likelihood function is intractable. 


\subsection{Cell Model}
The cell model~\citep{Vo2015} describes the collective spreading of
cells on a scratch assay, driven by the motility and proliferation of
individual cells, with particular applications in wound
healing~\citep[e.g.][]{Dale1994} and tumor
growth~\citep[e.g.][]{Swanson2003}. In the context of our work, the
experimental design is about deciding when to count the number of
cells on the scratch assay in order to optimally estimate the cell
diffusivity and proliferation rate.~\citet{Price2018b} used the Cell
Model before to compare the synthetic likelihood and approximate
Bayesian computation (ABC) likelihood-free inference approaches to
estimate these model parameters. Importantly, in their work they
assumed that they had access to $144$ images of a scratch assay and
went on to estimate the model parameters given that an experimenter
could analyse and quantify the cell spreading in all $144$ images. We
here wish to find out which of these images an experimenter should
analyse if there is a limited experimental budget.

The (discrete) cell model starts with a grid of size $27\times36$ and
$110$ initial cells that are randomly placed in the upper part of the
grid. This simulates wound healing, where a part of the tissue was
scratched away due to an accident. At each discrete time step, every
cell in the grid has a chance of moving to a neighbouring, empty grid
position, which is given by the model diffusivity $D$. Similarly, at
each discrete time every cell also has a chance to reproduce and spawn
a new cell in a neighbouring, empty position, which is dictated by the
model proliferation rate $\lambda$. While the model parameters of
interest are the diffusivity $D$ and proliferation rate $\lambda$, it
is often easier to work with the probability of motility $P_m \in
[0,1]$ and probability of proliferation $P_p \in [0,1]$.\footnote{See
  how these parameters can be converted in~\citet{Vo2015}.} For a
particular combination of $\{P_m, P_p\}$ we can then simulate a
time-series of grids where cells move around and
reproduce.\footnote{See Appendix~\ref{app:D} for a plot showing the
  spreading of cells under this model.} In the context of BED, the
discrete design variable is then the time at which to observe this
grid and count the total number of cells. In reality, a human would
have to physically count the number of cells under a microscope, which
is time-consuming, and therefore we want to find the optimal times at
which to have the experimenter make an observation. Similar to previous models, we here assume that we have access to a new, independent stochastic process for each sequential iteration.

We shall use 144 time steps as in~\citet{Vo2015}
and~\citet{Price2018b}, which means that, including the initial grid,
there are 145 grids in every time-series. For the summary statistics
used to compute~\eqref{eq:lfire_ratio}, we use the Hamming distance
between a particular grid and the initial grid, as well as the total
number of cells in a particular grid. The one-dimensional design
variable is discrete and can take values between 1 and 145, i.e.~$d
\in \{1, \dots, 145\}$, while the summary statistic is
two-dimensional. For the model parameters we use prior distributions
$p(P_m) = \mathcal{U}(P_m; 0, 1)$ and $p(P_p) = \mathcal{U}(P_p; 0,
0.005)$; we choose the true model parameters to be $P_{m, \text{true}}
= 0.35$ and $P_{p, \text{true}} = 0.001$ as
\citet{Price2018b}. Because the simulation time for the cell model is
significantly more expensive than the previous models we have tested,
we only use 300 initial prior samples during the sequential BED
algorithm, which we run up to five iterations. We note that, while
decreasing the computational resources needed, this may increase the
Monte-Carlo error in~\eqref{eq:sequtility_mc} and the error in the
LFIRE ratio estimate.
%
%

%
\begin{figure}[!t]
\includegraphics[width=\linewidth]{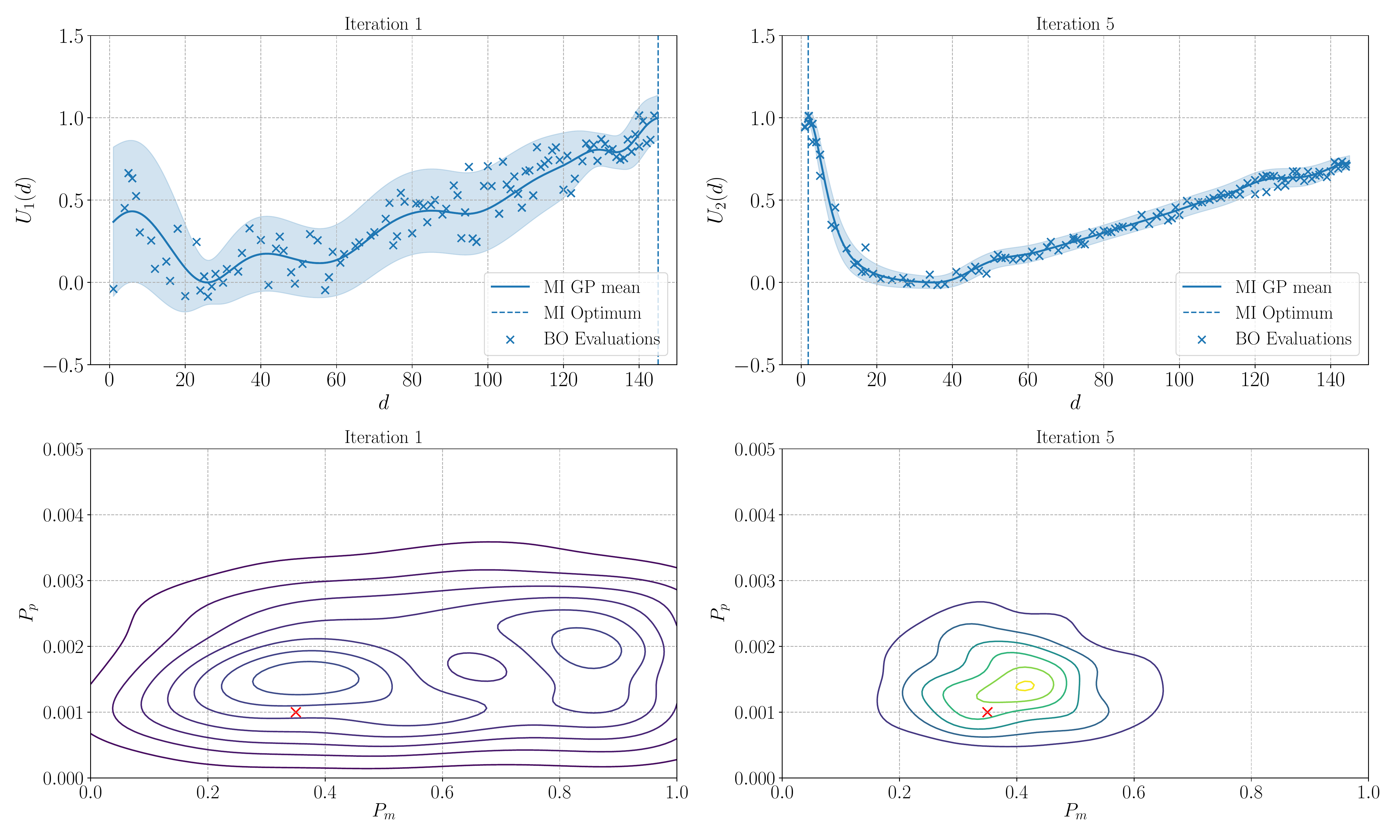}
\caption[]{MI utilities (top row) and posteriors (bottom row) for
  iteration one and five for the Cell model. The true model parameters are shown with a red cross. }
\label{fig:cell_comp}
\end{figure}

In the top row in Figure~\ref{fig:cell_comp} we compare the sequential
MI utilities for iteration 1 (left) and 5 (right) for the cell model;
see Appendix~\ref{app:G} for a plot showing utilities for all
iterations. Shown are the posterior predictive means and variances of
the surrogate GP model, the BO evaluations and the respective
optima. Because of the discrete domain, a normal GP tends to overfit
this data and therefore we have used a one-layer deep
GP~\citep[see][]{Damianou2013} as the surrogate model, which means
that the GP hyper-parameters are modelled by GPs as well. For each
iteration there seems to be some merit in taking observations at small
and large designs but not for medium-large designs, e.g.~$d \sim
20-40$. At early designs, not much proliferation will have happened
(due to its small prior probability) and so one can more easily
measure the effect of motility. Conversely, at large designs one can
average out the effect of motility and more easily notice the effect
of proliferation, as more time has elapsed. In our case, the
information gain about proliferation seems to beat that about motility
in iteration $1$, and similarly for iterations $2-4$ (see
Appendix~\ref{app:G}). After having repeatedly made measurements at
large designs, at iteration $5$ it becomes more effective to measure
at small designs. This is because we have decreased the uncertainty in
the proliferation parameter in iterations $1-4$ and then need a
measurement at early designs to sufficiently decrease the uncertainty
in the motility parameter. Note that we resampled the parameters
according to Algorithm~\ref{algo:resampling} before iteration $2$ and
$3$, as the effective sample size went below 50\%.

Similarly to before, we use KDE and updated belief samples obtained
from Algorithm~\ref{algo:postsamples} to visualise the approximate
posterior densities after every iteration. In the bottom row in
Figure~\ref{fig:cell_comp} we show the posterior densities obtained
after iteration $1$ (left) and $5$ (right); see Appendix~\ref{app:G}
for a plot showing posterior distributions after every
iteration. After iteration $1$, the updated belief distribution has a
wide spread in the $P_m$ parameter and is more narrow for the $P_p$
parameter. The optimal design at iteration $1$ was at the far end of
the design domain at $d^\ast_1 = 145$. This is a design that helps to
more easily detect the effect of proliferation as opposed to motility,
which is reflected in the figure. The same phenomenon occurs in
subsequent iterations $2-4$, where the optimal designs are at the far
end of the design domain (see Appendix~\ref{app:G}). This results in
posterior distributions that are relatively narrow for $P_p$ but wider
for $P_m$. Taking a measurement at the small design $d_5^\ast=2$ in
iteration $5$, which allows us to more easily detect the effect of
motility, reduces the uncertainty in the $P_m$ parameter as well, as
can be seen in the bottom right plot of
Figure~\ref{fig:cell_comp}. The mode of the posterior distribution
after iteration $5$ is close to the true parameter value of $P_{m,
  \text{true}} = 0.35$ and $P_{p, \text{true}} = 0.001$. The estimated
mean of the motility parameter is $\widehat{P}_{m}=0.394$ with a
$95\%$ credibility interval of $[0.166, 0.642]$, while for the
proliferation parameter it is $\widehat{P}_{p}=0.00150$ with a $95\%$
credibility interval of $[0.00055, 0.00265]$. Both credibility
intervals contain the true parameter values. The credibility intervals
were computed using a Gaussian KDE of the marginal posterior samples
and the HPDI method.

The cell model demonstrates that we can effectively use MI in
sequential BED when the forward simulations are expensive and the
design domain is discrete. For this model, an experimenter might
intuitively want to take observations at regular intervals but we have
seen from the sequential utility functions that the expected
information gain may then be sub-optimal. Sequential BED suggests
that, when there is a limited budget, it is best to first take
observations at high values in the design domain as the effect of
proliferation dominates that of motility. Later, however, we should
take observations at small designs to reduce the uncertainty in the
motility parameter as well.


\section{Conclusion} \label{sec:conclusion}

In this work we have presented a sequential BED framework for implicit
models, where the data-generating distribution is intractable but
sampling from it is possible. Our framework uses the mutual
information (MI) between model parameters and data as the utility
function, which has not been done before in the context of sequential
BED for implicit models due to computational difficulties. In
particular, we showed how to obtain an estimate of the MI by density
ratio estimation methods and then optimise it with Bayesian
optimisation. To estimate the MI in subsequent iterations, we
showed how to obtain updated belief samples by using a weighted
particle approach and updating weights every iteration using the
computed density ratios. We devised a resampling algorithm that yields
new parameter samples whenever the effective sample size of these
weights went below a minimum threshold. The framework can be used to
produce sequential optimal experimental designs that can guide the
data-gathering phase in a scientific experiment.

We first illustrated and explained our framework on a oscillatory toy
model with multi-modal posteriors and then applied it to more
challenging examples from epidemiology and cell spreading. For all
examples we obtained optimal experimental designs that made intuitive
sense and and resulted in informative posterior distributions. For the
oscillatory toy model 
Bayesian D-Optimality (BD-Opt), which has been used in sequential BED
once before by~\citet{Hainy2016b}. We found that, besides being less
computationally expensive, MI usually led to different optimal designs
than BD-Opt, due to the latter penalising multi-modality and only
focussing on posterior precision.

While we have applied our framework to implicit models with low
dimensionality, the theory is general and extends to models with
high-dimensional designs as well, albeit being more computationally
intensive. Standard Bayesian optimisation, as used in this work,
becomes, however, expensive and less effective in high dimensions. One
would either have to utilise recent advances in high-dimensional
Bayesian optimisation or look towards alternative gradient-free
optimisation schemes, such as for example approximate coordinate exchange
~\citep{Overstall2017}.


The MI utility represents the information gain of an experiment and is
thus focused on obtaining accurate estimation results. However, it
does not take the computational or financial cost of the different
experimental designs into account. For that purpose one may want to
maximise a normalised information gain instead, where we for example
divide the MI by the estimated cost of running the experiment.
In our paper, we focused on experimental design for parameter
estimation. However, we note that the proposed framework could also be
applied to experimental design for model discrimination, as well as
dual-purpose model discrimination and parameter estimation, by
conditioning the data-generating process on a particular model and
then averaging over the model space.

%% file: supplementary.tex
\section{Alternative form of the BD-Opt Utility} \label{app:A}

We noticed certain numerical instabilities with the Bayesian D-Optimality (BD-Opt) utility as used by~\citet{Hainy2016b},
\begin{equation} \label{eq:bdopt2}
U(\dbf) = \mathbb{E}_{p(\ybf \mid \dbf)}\left[\frac{1}{\text{det}(\text{cov}(\thetab \mid \ybf, \dbf))}\right].
\end{equation}
These instabilities arise because inside the above expectation we are computing the inverse of the determinant of the posterior covariance. If the exact value of ${\text{det}(\text{cov}(\thetab \mid \ybf, \dbf))}$ is small, approximating the expectation in~\ref{eq:bdopt2} with a standard sample-average may lead to extremely large $U(\dbf)$ evaluations. Furthermore, for implicit models we cannot compute the determinant of the covariance exactly but have to approximate it with samples from the posterior distribution, obtained via a sequential Monte-Carlo approach~\citep[see][]{Hainy2016b}). Poor approximations of this quantity may also lead to large spikes in utility evaluations.
We partly rectified this in our approach by taking the median instead of the mean in the sampling-based computation of the expectation. 

Ultimately, these spikes in utility evaluations arise from an inherent instability in the BD-Opt utility. Although we have not tested it, we believe that a more stable form of the BD-Opt utility might be the following,
\begin{equation} \label{eq:bdopt2-stable}
U_{\text{stable}}(\dbf) = -\mathbb{E}_{p(\ybf \mid \dbf)}\left[\log\left\{\text{det}(\text{cov}(\thetab \mid \ybf, \dbf))\right\}\right].
\end{equation}
The natural logarithm of the determinant of the posterior is additively proportional to the differential entropy of the multivariate normal distribution. Thus, similar to the previous BD-Opt, this utility works well for posterior distributions that a nearly Gaussian and fails for highly non-Gaussian posteriors, e.g.~multi-modal distributions. Furthermore, by applying Jensen's inequality for concave functions, the utility $U_{\text{stable}}(\dbf)$ in~\ref{eq:bdopt2-stable} can be interpreted as a lower bound on the logarithm of $U(\dbf)$ in~\ref{eq:bdopt2}.

\section{Utility estimation with weighted samples} \label{app:B}

We could also approximate the sequential mutual information utility in~\eqref{eq:sequtility_implicit} directly with weighted prior samples instead of using posterior samples, i.e~
\begin{align}
\widehat{U}_k(\dbf) =& \frac{1}{N} \sum_{i=1}^N \log\left[\widehat{r}_k(\dbf, \ybf^{(i)}, \thetab^{(i)}, \mathbb{D}_{k-1})\right] w_{k-1}(\thetab^{(i)}; \mathbb{D}_{k-1}) , \label{eq:sequtility_mc_app}
\end{align}
where $\ybf^{(i)} \sim p(\ybf \mid \dbf, \thetab^{(i)})$, $\thetab^{(i)} \sim p(\thetab)$ and the weights $w_{k-1}$ are given by
\begin{align}
w_{k-1}(\thetab; \mathbb{D}_{k-1}) = \prod_{s=1}^{k-1} \widehat{r}_{s}(\dbf_{s}^\ast, \ybf_{s}^\ast, \thetab, \mathbb{D}_{s-1}), \label{eq:weights_app}
\end{align}
with $\widehat{r}_{1}(\dbf_{1}^\ast, \ybf_{1}^\ast, \thetab, \mathbb{D}_{0}) = \widehat{r}_{1}(\dbf_{1}^\ast, \ybf_{1}^\ast, \thetab)$ according to~\eqref{eq:ratio} and $w_0(\thetab) = 1 \, \forall \, \thetab$.

In theory, the estimator in~\eqref{eq:sequtility_mc_app} has a lower variance than the estimator in~\eqref{eq:sequtility_mc} that we have used. In practice, however, we did not observe a significant difference but noticed that the estimator in~\eqref{eq:sequtility_mc_app} had longer computation times. Thus, we opted to use the sampling-based approach in~\eqref{eq:sequtility_mc} instead.

\section{Parameter Space Transformation} \label{app:C}

We here describe in more detail how to transform the parameter samples $\thetab$ and boundary conditions $\mathcal{B}$ before the resampling procedure explained in Section~\ref{sec:resampling}. Let $\theta_j^{(i)}$ be the jth element of the parameter sample $\thetab^{(i)}$. If the parameters $\theta_j$ in $\thetab$ have different scales, the KD-Tree algorithm produces nearest neighbours that underestimate, or overstimate, the standard deviation $\sigma$ during the resampling procedure (see section~\ref{sec:resampling}). We found that we can overcome this and increase robustness by transforming all parameter samples such that their elements are bound between $0$ and $1$. Thus, for every element of every parameter sample we do the transformation $\theta_j^{\prime (i)} \leftarrow \theta_j^{(i)}$ as follows,
\begin{equation} \label{eq:transf}
\theta_j^{\prime (i)} = \frac{\theta_j^{(i)} - \theta_j^{\text{min}}}{\theta_j^{\text{max}} - \theta_j^{\text{min}}},
\end{equation}
where $\theta_j^{\text{max}}$ and $\theta_j^{\text{min}}$ are the maximum and minimum, respectively, of the set of parameter samples $\{\theta_j^{(i)}\}_{i=1}^N$ for the jth element. Now consider the boundary conditions $\mathcal{B}_{j} = [\mathcal{B}_{j}^{-}, \mathcal{B}_{j}^{+}]^\top$ for the jth element of the parameter $\thetab$, where $\mathcal{B}_{j}^{-}$ and $\mathcal{B}_{j}^{+}$ are the lower and upper boundary, respectively. We assume that beyond these boundaries the prior probability $p(\theta_j)$ is zero and therefore we cannot resample beyond these boundaries. Using the same $\theta_j^{\text{max}}$ and $\theta_j^{\text{min}}$ as before, we transform the boundaries as well, i.e.
\begin{equation}
\mathcal{B}_{j}^{\prime -/+} = \frac{\mathcal{B}_{j}^{-/+} - \theta_j^{\text{min}}}{\theta_j^{\text{max}} - \theta_j^{\text{min}}}.
\end{equation}
In order to transform the resampled parameter samples back to the original parameter space, we simply have to invert~\eqref{eq:transf} and obtain an expression for $\theta_j^{(i)}$.  


\section{Simulation Plots for all Models} \label{app:D}

\begin{figure}[!t]
    \begin{minipage}[b]{0.48\textwidth}
    	\centering
        \includegraphics[width=1\linewidth]{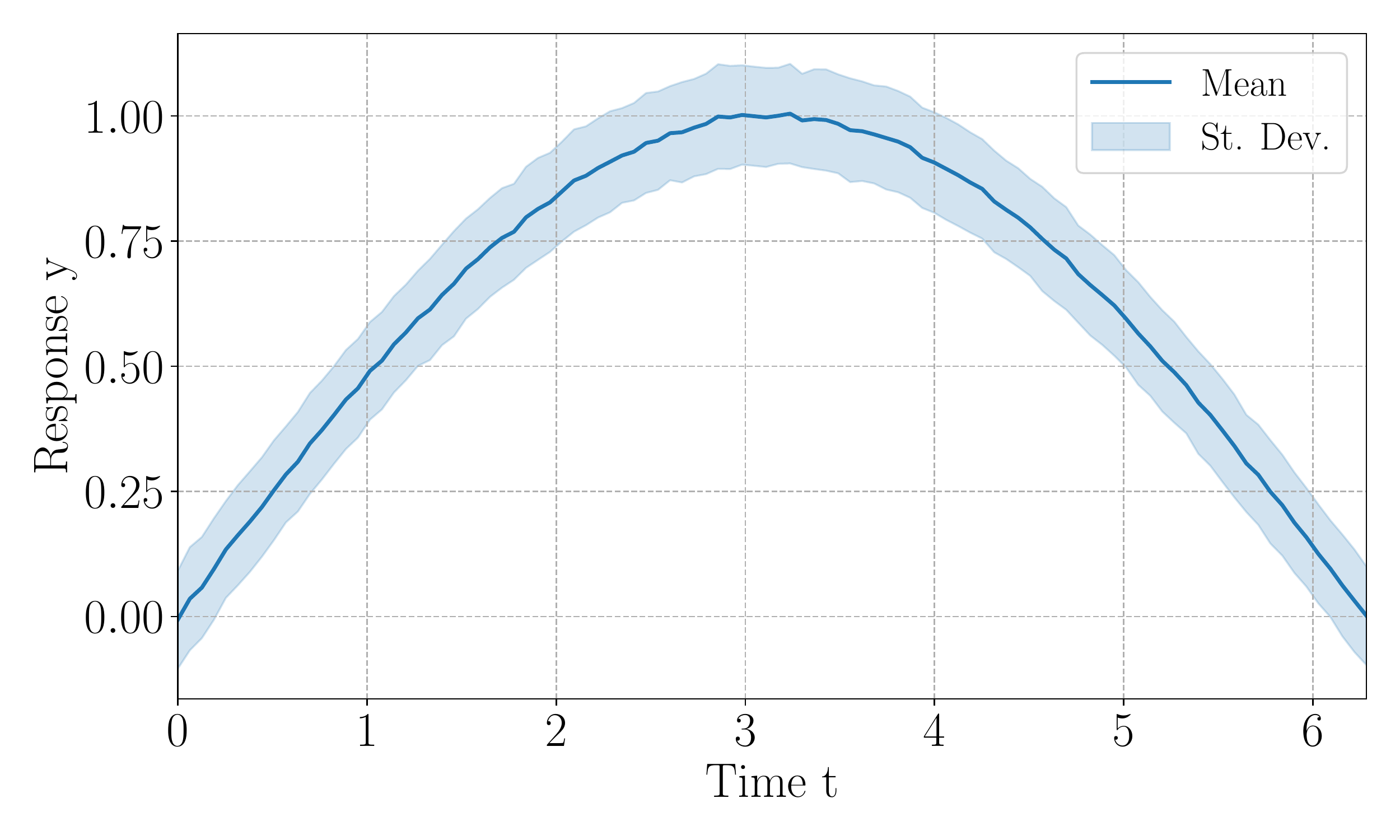}
        \caption[]{Sine model response as a function of time, computed with the true model parameters.}
        \label{fig:sine_sim}
    \end{minipage}
    \hfill
    \begin{minipage}[b]{0.48\textwidth}
    	\centering
        \includegraphics[width=1\linewidth]{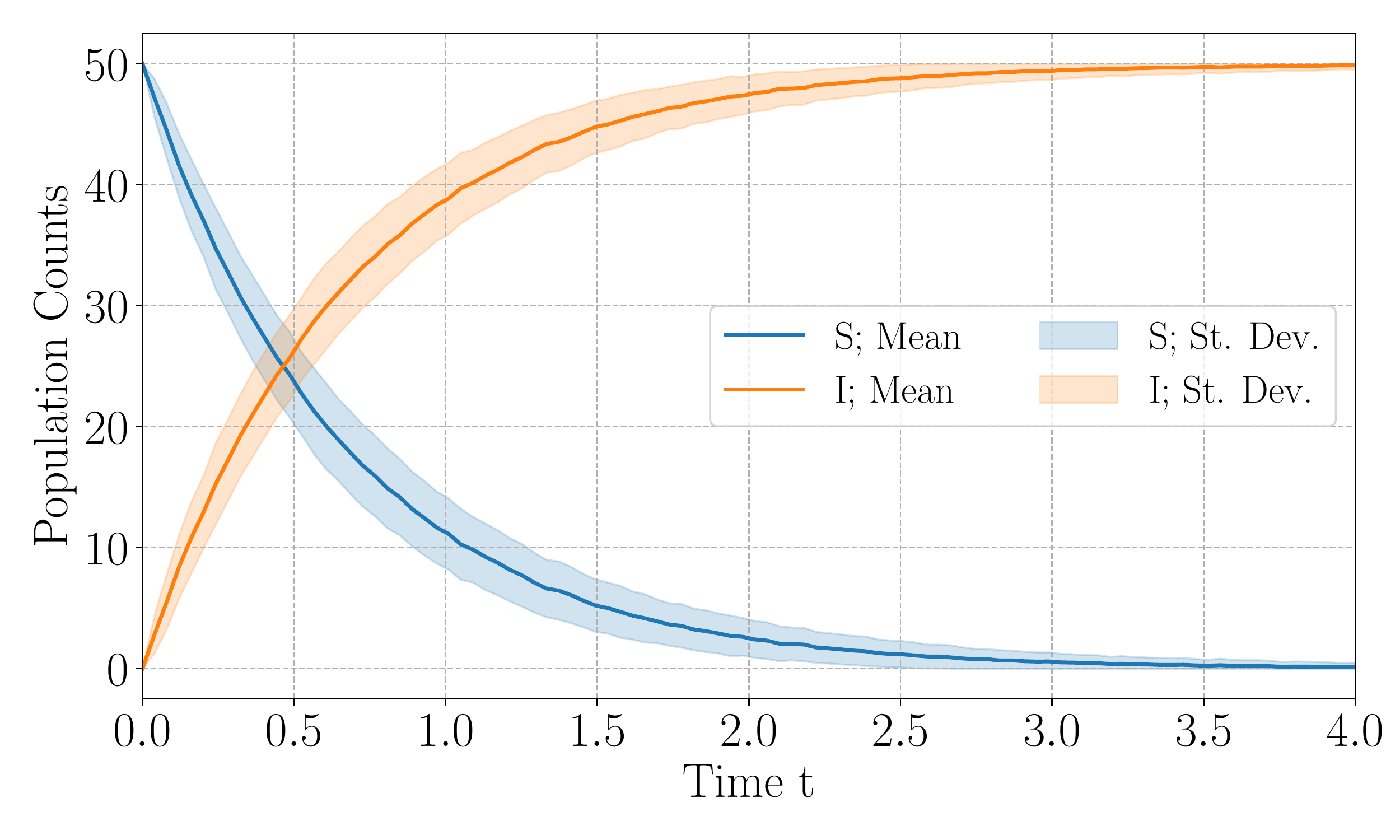}
        \caption[]{Death model population counts of S and I as a function of time, computed with the true model parameters.}
        \label{fig:death_sim}
    \end{minipage}
    \hfill
    \begin{minipage}[b]{0.48\textwidth}
    	\centering
        \includegraphics[width=1\linewidth]{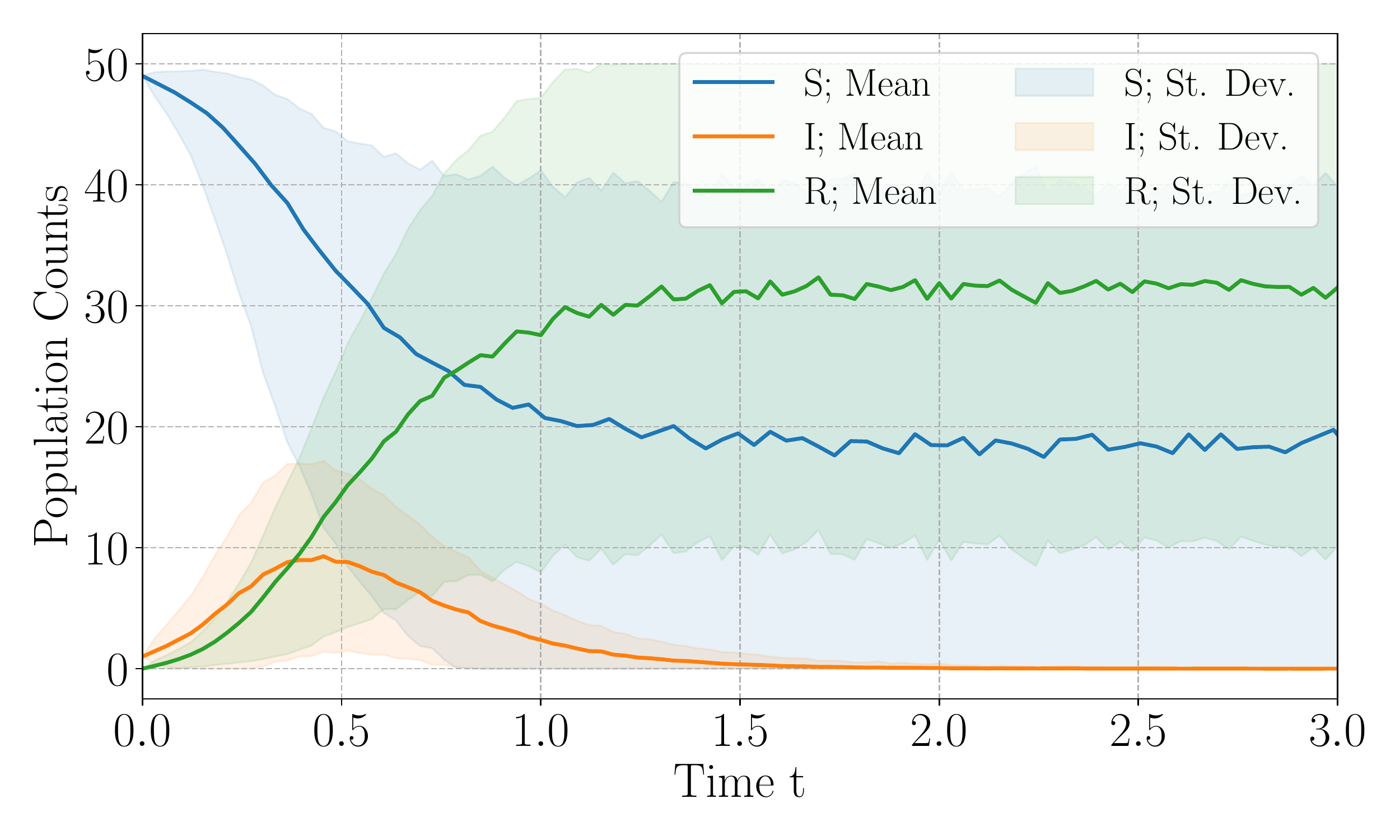}
        \caption[]{SIR model population counts of S, I and R as a function of time, computed with the true model parameters.}
        \label{fig:sir_sim}
    \end{minipage}
    \hfill
    \begin{minipage}[b]{0.48\textwidth}
    	\centering
        \includegraphics[width=1\linewidth]{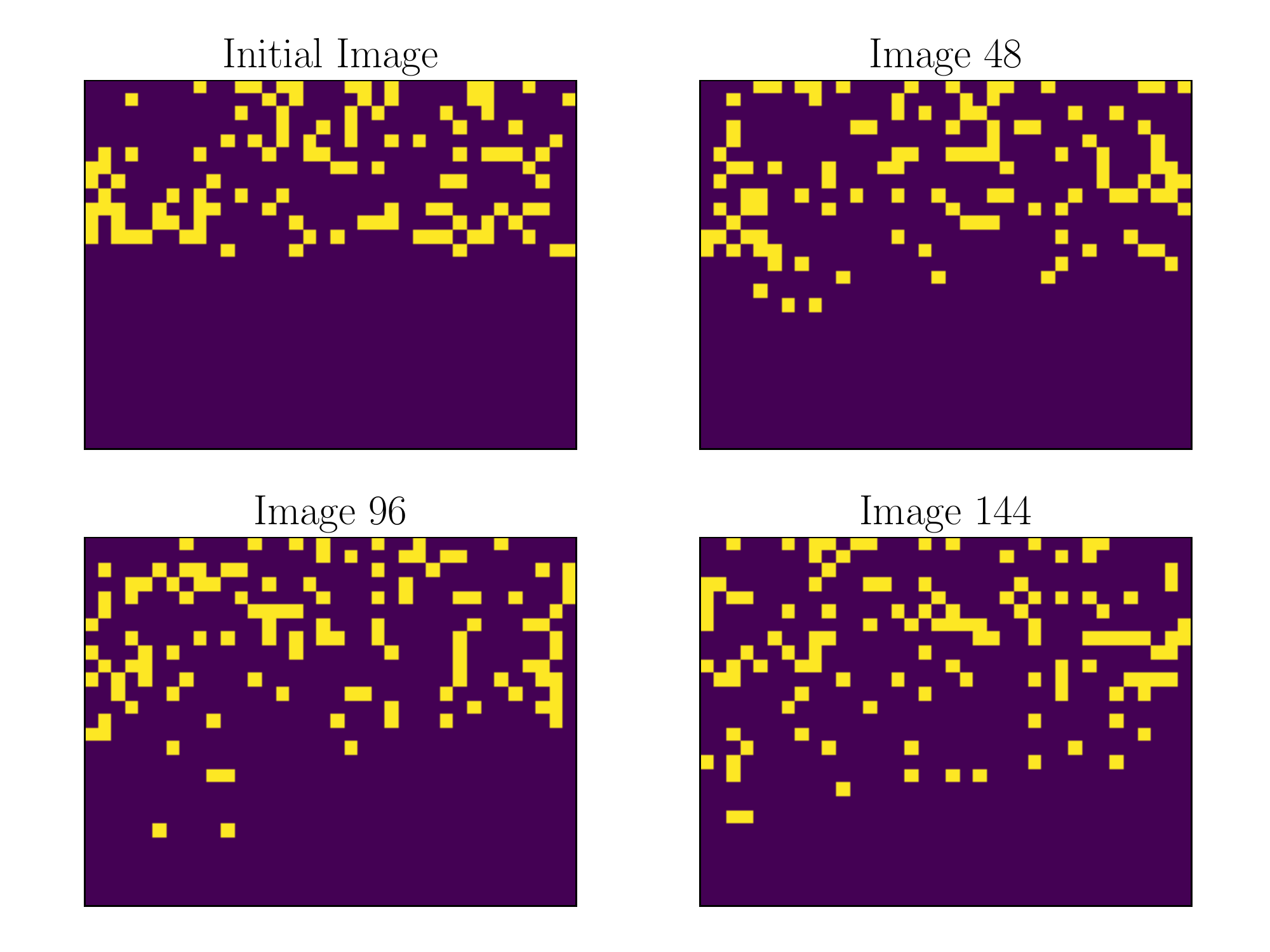}
        \caption[]{Cell model example simulation of cell motility and proliferation as a function of time, computed with the true model parameters.}
        \label{fig:cell_sim}
    \end{minipage}
\end{figure}

We show simulations of data as a function of time for all models considered in the main text in Figure~\ref{fig:sine_sim} (oscillation toy model), Figure~\ref{fig:death_sim} (death model), Figure~\ref{fig:sir_sim} (SIR model) and Figure~\ref{fig:cell_sim} (cell model). For the oscillation toy model, death model and SIR model we show means and standard deviations computed from $1{,}000$ simulations of time-series. For the cell model we show images of the spread of cells at different timesteps between $1$ and $144$. For each model, all responses were simulated using the corresponding true model parameters considered in the main text.

\pagebreak 

\section{Reference MI Computation} \label{app:E}

In order to compute reference mutual information (MI) values for the oscillation toy model and the death model, we use a nested Monte-Carlo sample-average. Note that we only compute reference MI values for the first sequential iteration. We assume that we can readily evaluate the data-generation distribution $p(\ybf \mid \thetab, \dbf)$ for these models and use this to compute a sample-average of the marginal data distribution, i.e.~$p(\ybf \mid \dbf) \approx \frac{1}{M}\sum_{j=1}^M p(\ybf \mid \thetab^{(j)}, \dbf)$, where $\thetab^{(j)} \sim p(\thetab)$. The mutual information is then approximated by
\begin{align}
  \mathrm{I}(\thetab; \ybf \mid \dbf) 
  &= \int p(\thetab, \ybf \mid \dbf) \log\left[\frac{p(\thetab, \ybf \mid \dbf)}{p(\thetab)p(\ybf \mid \dbf)}\right] \mathrm{d}\thetab \mathrm{d}\ybf \\
  &= \int p(\ybf \mid \thetab, \dbf) p(\thetab) \log\left[\frac{p(\ybf \mid \thetab, \dbf)}{p(\ybf \mid \dbf)}\right] \mathrm{d}\thetab \mathrm{d}\ybf \\
  &\approx \frac{1}{N} \sum_{i=1}^N \log\left[\frac{p(\ybf^{(i)} \mid \thetab^{(i)}, \dbf)}{\frac{1}{M}\sum_{j=1}^M p(\ybf^{(i)} \mid \thetab^{(j)}, \dbf)}\right], \label{eq:reference_mi}
\end{align}
where $\ybf^{(i)} \sim p(\ybf \mid \dbf, \thetab^{(i)})$, $\thetab^{(i)} \sim p(\thetab)$ and $\thetab^{(j)} \sim p(\thetab)$. 

For the sine model we use $p(y \mid \omega, t) = \mathcal{N}(y; \sin(\omega t), 0.1^2)$, i.e.~\eqref{eq:sinean}, in order to compute the reference MI according to~\eqref{eq:reference_mi} with $N=M=1{,}000$. For the death model we use the data-generating distribution $p(S \mid b, \tau) = \mathrm{Bin}(S; S_0, \exp(-b(\tau - \tau_0)))$, where $S$ is the number of susceptible individuals, $S_0 = 50$ and $\tau_0 = 0$~\citep{Cook2008, Kleinegesse2019}. The number of susceptibles can be computed from the number of infected individuals $I$ by $S=S_0 - I$. We then compute the reference MI using~\eqref{eq:reference_mi} and $N=M=1{,}000$.

\section{Additional plots for the Death Model} \label{app:F}

\begin{figure}[!h]
\includegraphics[width=\linewidth]{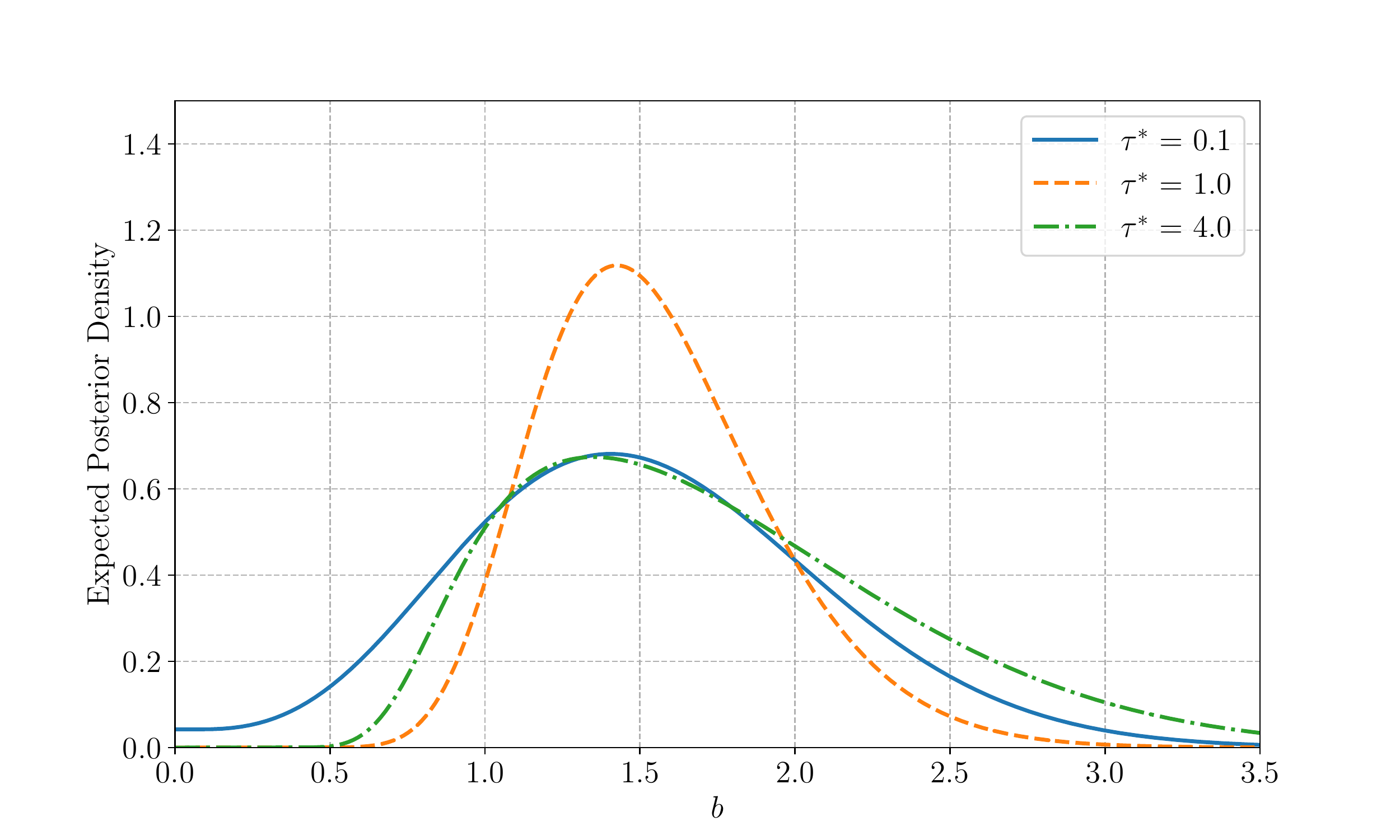}
\caption[]{Expected posterior densities for the Death model at different measurement times $\tau^\ast \in \{0.1, 1.0, 4.0\}$, averaged over $100$ observations $I(\tau^\ast)$.}
\label{fig:death_post_diffloc}
\end{figure}

In Figure~\ref{fig:death_post_diffloc} we show average posterior densities at different measurement times for the Death model. The posterior densities were approximated by using the analytic likelihood of the model and averaged over $100$ observations $I(\tau^\ast)$ at $\tau^\ast \in \{0.1, 1.0, 4.0\}$.

\begin{figure}[!h]
\includegraphics[width=\linewidth]{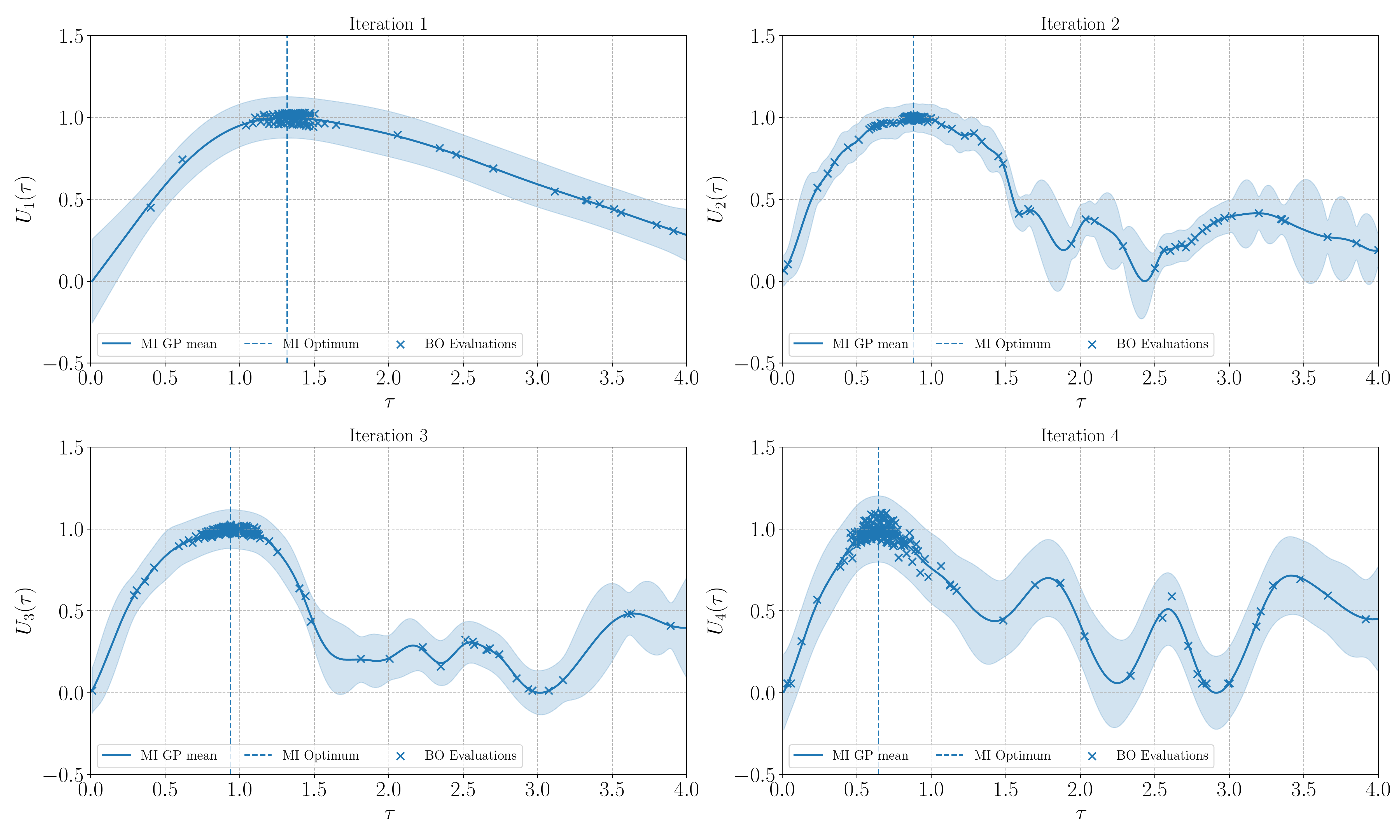}
\caption[]{Sequential MI utilities for all iterations of the Death model. Shown are the GP mean and variance, BO evaluations and optimum of the GP mean}
\label{fig:death_utils_bo}
\end{figure}

In Figure~\ref{fig:death_utils_bo} we show the sequential MI utilities for all iterations of the Death model, including the GP means and variances.

\section{Additional plots for the Cell model} \label{app:G}

\begin{figure}[!t]
\includegraphics[width=\linewidth]{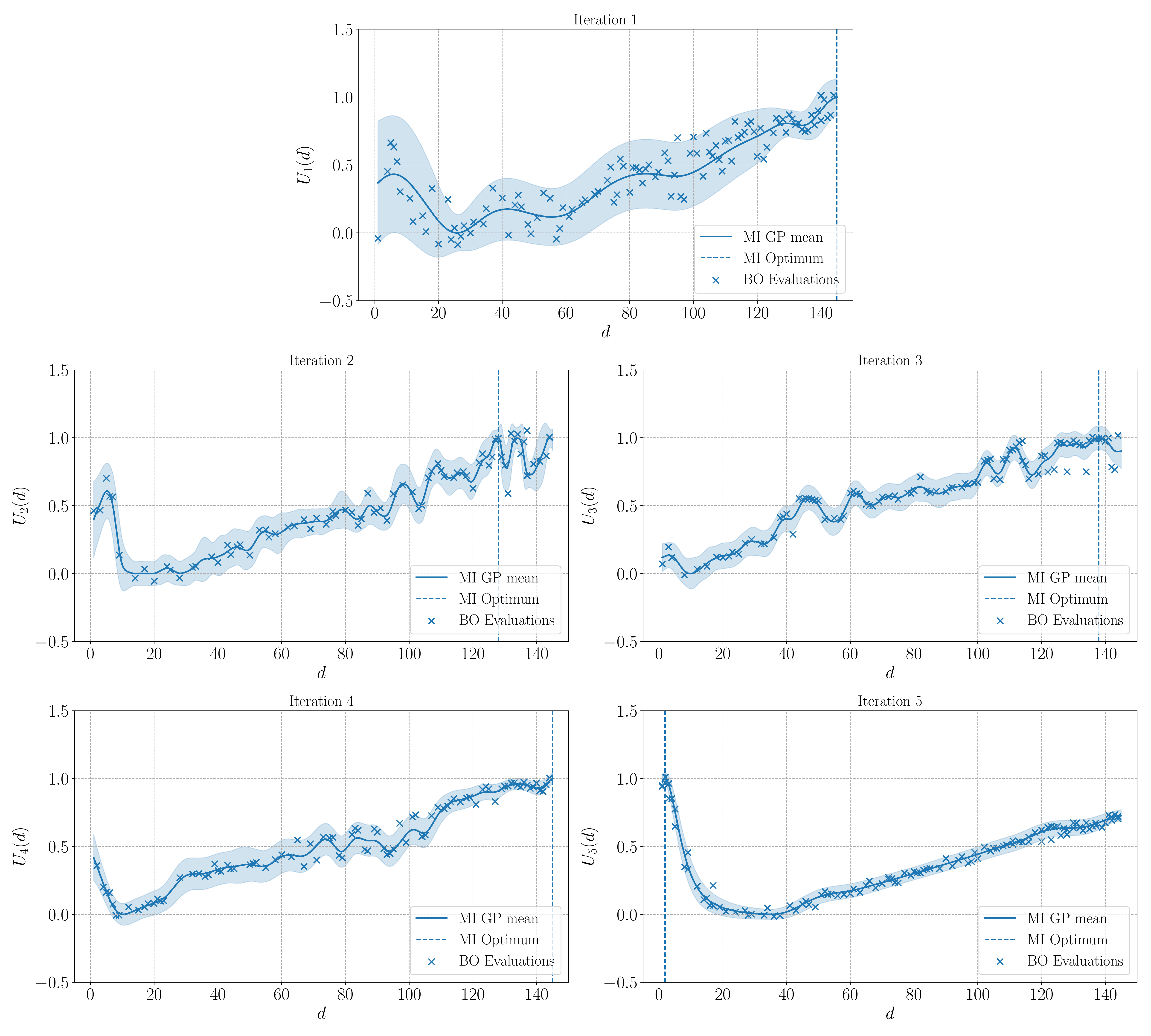}
\caption[]{MI utilities at different iterations for the Cell model. Shown are the GP means and variances, BO evaluations and optima of the GP means.}
\label{fig:cell_utils}
\end{figure}

\begin{figure}[!t]
\includegraphics[width=\linewidth]{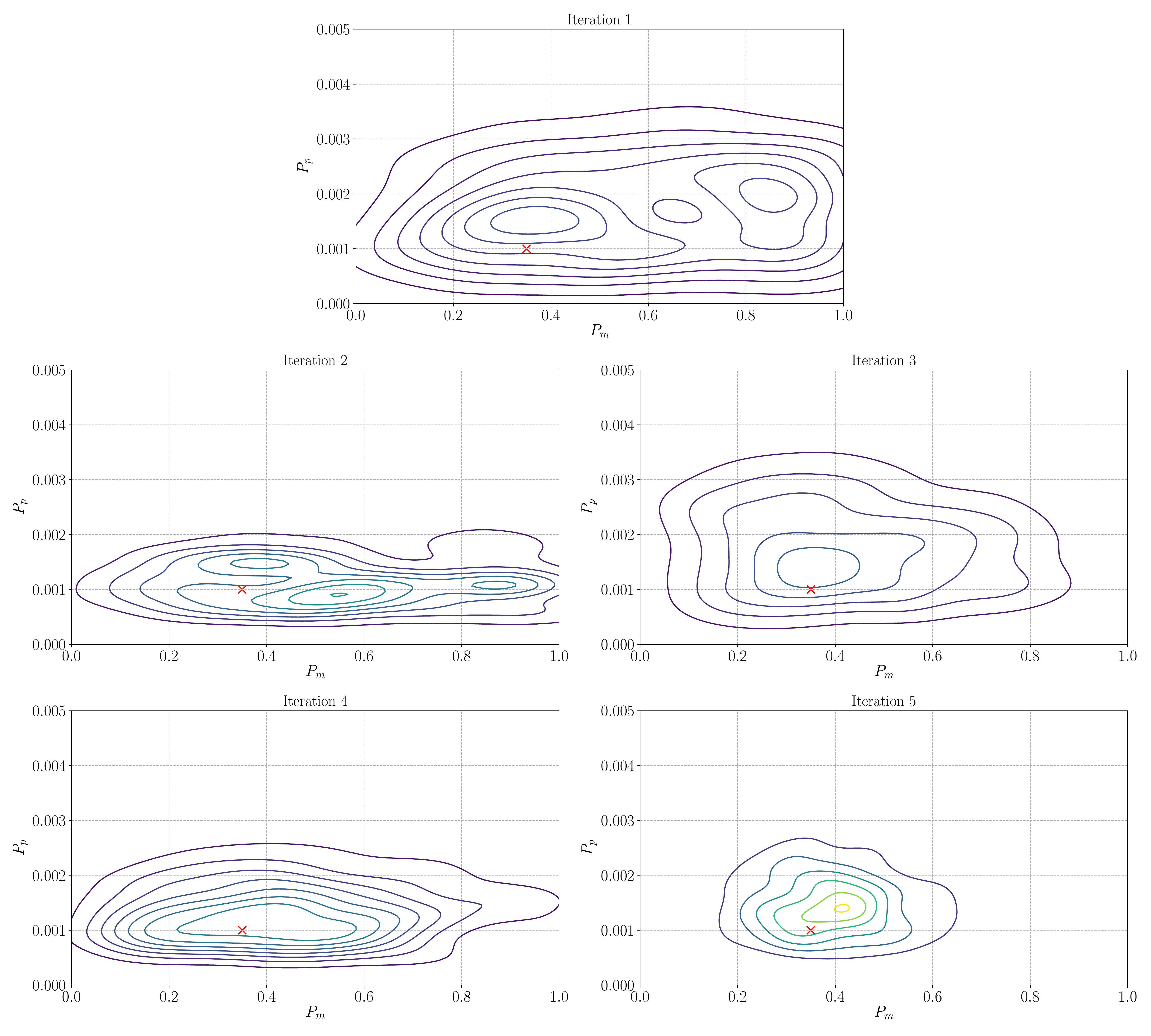}
\caption[]{Comparison of updated belief distribution densities at different iterations for the Cell model.}
\label{fig:cell_posts}
\end{figure}

In Figure~\ref{fig:cell_utils} we show the MI utilities for every iteration for the Cell model. We show the corresponding posterior distributions after every iteration in Figure~\ref{fig:cell_posts}.

